\documentclass{article}
\usepackage{ijcai23}

\usepackage{times}
\usepackage{soul}
\usepackage{url}
\usepackage[hidelinks]{hyperref}
\usepackage[utf8]{inputenc}
\usepackage[small]{caption}
\usepackage{graphicx}
\usepackage{amsmath}
\usepackage{amsthm}
\usepackage{booktabs}
\usepackage{algorithmic}
\usepackage[switch]{lineno}

\usepackage{color}
\usepackage{subcaption}
\usepackage[hidelinks]{hyperref}

\usepackage{amsmath,amsfonts,bm, bbm}
\usepackage{xparse}

\DeclareDocumentCommand\W{ g g }{%
        \IfNoValueTF {#1} {\mathbf{W}} {
            \IfNoValueTF {#2} {\mathbf{W}^{(#1)}}{\mathbf{W}^{(#1)}_{#2}}
        }
}

\DeclareDocumentCommand\bias{ g g }{%
        \IfNoValueTF {#1} {\mathbf{b}} {
            \IfNoValueTF {#2} {\mathbf{b}^{(#1)}}{\mathbf{b}^{(#1)}_{#2}}
        }
}

\DeclareDocumentCommand\betavar{ g g }{%
        \IfNoValueTF {#1} {\bm{\beta}} {
            \IfNoValueTF {#2} {{\bm{\beta}^{(#1)}}{}}{\bm{\beta}^{(#1)}_{#2}}
        }
}

\DeclareDocumentCommand\xivar{ g g }{%
        \IfNoValueTF {#1} {\bm{\xi}} {
            \IfNoValueTF {#2} {{\bm{\xi}^{(#1)}}{}}{\bm{\xi}^{(#1)}_{#2}}
        }
}

\DeclareDocumentCommand\xivarn{ g g }{%
        \IfNoValueTF {#1} {\bm{\xi^-}} {
            \IfNoValueTF {#2} {\bm{\xi^-}^{+(#1)}}{\bm{\xi^-}^{+(#1)}_{#2}}
        }
}

\DeclareDocumentCommand\xivarp{ g g }{%
        \IfNoValueTF {#1} {\bm{\xi^+}} {
            \IfNoValueTF {#2} {\bm{\xi^+}^{+(#1)}}{\bm{\xi^+}^{+(#1)}_{#2}}
        }
}

\DeclareDocumentCommand\nuvar{ g g }{%
        \IfNoValueTF {#1} {\bm{\nu}} {
            \IfNoValueTF {#2} {{\bm{\nu}^{(#1)}}{}}{\bm{\nu}^{(#1)}_{#2}{}}
        }
}

\DeclareDocumentCommand\hnuvar{ g g }{%
        \IfNoValueTF {#1} {\bm{\hat{\nu}}} {
            \IfNoValueTF {#2} {{\bm{\hat{\nu}}^{(#1)}}{}}{\bm{\hat{\nu}}^{(#1)}_{#2}{}}
        }
}

\DeclareDocumentCommand\muvar{ g g }{%
        \IfNoValueTF {#1} {\bm{\mu}} {
            \IfNoValueTF {#2} {{\bm{\mu}^{(#1)}}{}}{\bm{\mu}^{(#1)}_{#2}}
        }
}

\DeclareDocumentCommand\gammavar{ g g }{%
        \IfNoValueTF {#1} {\bm{\gamma}} {
            \IfNoValueTF {#2} {{\bm{\gamma}^{(#1)}}{}}{\bm{\gamma}^{(#1)}_{#2}}
        }
}

\DeclareDocumentCommand\lambdavar{ g g }{%
        \IfNoValueTF {#1} {\bm{\lambda}} {
            \IfNoValueTF {#2} {{\bm{\lambda}^{(#1)}}{}}{\bm{\lambda}^{(#1)}_{#2}}
        }
}

\DeclareDocumentCommand\tbetavar{ g g }{%
        \IfNoValueTF {#1} {{\bm{\tilde{\beta}}}} {
            \IfNoValueTF {#2} {{{\bm{\tilde{\beta}}}^{(#1)}}{}}{{{\bm{\tilde{\beta}}}^{(#1)}_{#2}}}
        }
}

\DeclareDocumentCommand\alphavar{ g g }{%
        \IfNoValueTF {#1} {\bm{\alpha}} {
            \IfNoValueTF {#2} {{\bm{\alpha}^{(#1)}}}{\bm{\alpha}^{(#1)}_{#2}}
        }
}

\DeclareDocumentCommand\D{ g g }{%
        \IfNoValueTF {#1} {\mathbf{D}} {
            \IfNoValueTF {#2} {\mathbf{D}^{(#1)}}{\mathbf{D}^{(#1)}_{#2}}
        }
}

\DeclareDocumentCommand\A{ g g }{%
        \IfNoValueTF {#1} {\mathbf{A}} {
            \IfNoValueTF {#2} {\mathbf{A}^{(#1)}}{\mathbf{A}^{(#1)}_{#2}}
        }
}

\DeclareDocumentCommand\AA{ g g }{
        \IfNoValueTF {#1} {\mathbf{\Omega}} {
            \IfNoValueTF {#2} {\mathbf{\Omega}(#1, #1)}{\mathbf{\Omega}(#1, #2)}
        }
}

\DeclareDocumentCommand\S{ g g }{%
        \IfNoValueTF {#1} {\mathbf{S}} {
            \IfNoValueTF {#2} {\mathbf{S}^{(#1)}}{\mathbf{S}^{(#1)}_{#2}}
        }
}

\DeclareDocumentCommand\K{ g g }{%
        \IfNoValueTF {#1} {\mathbf{K}} {
            \IfNoValueTF {#2} {\mathbf{K}^{(#1)}}{\mathbf{K}^{(#1)}_{#2}}
        }
}

\DeclareDocumentCommand\B{ g g }{%
        \IfNoValueTF {#1} {\mathbf{B}} {
            \IfNoValueTF {#2} {\mathbf{B}^{(#1)}}{\mathbf{B}^{(#1)}_{#2}}
        }
}

\DeclareDocumentCommand\lowerb{ g g }{%
        \IfNoValueTF {#1} {{\mathbf{\underline{b}}}} {
            \IfNoValueTF {#2} {{\mathbf{\underline{b}}}^{(#1)}}{{\mathbf{\underline{b}}}^{(#1)}_{#2}}
        }
}

\DeclareDocumentCommand\z{ g g }{%
        \IfNoValueTF {#1} {z} {
            \IfNoValueTF {#2} {z^{(#1)}}{z^{(#1)}_{#2}}
        }
}

\DeclareDocumentCommand\s{ g g }{%
        \IfNoValueTF {#1} {s} {
            \IfNoValueTF {#2} {s^{(#1)}}{s^{(#1)}_{#2}}
        }
}

\DeclareDocumentCommand\dom{ g g }{%
        \IfNoValueTF {#1} {\mathcal{S}} {
            \IfNoValueTF {#2} {\mathcal{S}_{#1}}{\mathcal{S}^{#1}_{#2}}
        }
}

\DeclareDocumentCommand\domlb{ g g }{%
        \IfNoValueTF {#1} {\mathsf{LB}} {
            \IfNoValueTF {#2} {\mathsf{LB}(\mathcal{#1})}{\mathsf{LB}(\mathcal{#1}_{#2})}
        }
}

\DeclareDocumentCommand\domub{ g g }{%
        \IfNoValueTF {#1} {\mathsf{UB}} {
            \IfNoValueTF {#2} {\mathsf{UB}(\mathcal{#1})}{\mathsf{UB}(\mathcal{#1}_{#2})}
        }
}

\DeclareDocumentCommand\uns{ g g }{%
        \IfNoValueTF {#1} {\tilde{s}} {
            \IfNoValueTF {#2} {\tilde{s}_{#1}}{s^{(#1)}_{#2}}
        }
}

\DeclareDocumentCommand\ub{ g g }{%
        \IfNoValueTF {#1} {u} {
            \IfNoValueTF {#2} {u^{(#1)}}{u^{(#1)}_{#2}}
        }
}

\DeclareDocumentCommand\lb{ g g }{%
        \IfNoValueTF {#1} {l} {
            \IfNoValueTF {#2} {l^{(#1)}}{l^{(#1)}_{#2}}
        }
}

\DeclareDocumentCommand\hz{ g g }{%
        \IfNoValueTF {#1} {\hat{z}} {
            \IfNoValueTF {#2} {\hat{z}^{(#1)}}{\hat{z}^{(#1)}_{#2}}
        }
}

\DeclareDocumentCommand\bu{ g g }{%
        \IfNoValueTF {#1} {\mathbf{u}} {
            \IfNoValueTF {#2} {\mathbf{u}^{(#1)}}{\mathbf{u}^{(#1)}_{#2}}
        }
}

\DeclareDocumentCommand\bl{ g g }{%
        \IfNoValueTF {#1} {\mathbf{l}} {
            \IfNoValueTF {#2} {\mathbf{l}^{(#1)}}{\mathbf{l}^{(#1)}_{#2}}
        }
}

\DeclareDocumentCommand\aaa{ g }{%
        \IfNoValueTF {#1} {\bm{a}} {
            {\bm{a}^{({#1})}}
        }
}

\DeclareDocumentCommand\haaa{ g }{%
        \IfNoValueTF {#1} {\bm{\hat{a}}} {
            {\bm{\hat{a}}^{({#1})}}
        }
}

\DeclareDocumentCommand\bbb{ g g }{%
        \IfNoValueTF {#1} {\mathbf{P}} {
            \IfNoValueTF {#2} {{\mathbf{P}_{#1}}}{{\mathbf{P}_{#1}^{({#2})}}}
        }
}

\DeclareDocumentCommand\hbbb{ g g }{%
        \IfNoValueTF {#1} {\mathbf{\hat{P}}} {
            \IfNoValueTF {#2} {{\mathbf{\hat{P}}_{#1}}}{{\mathbf{\hat{P}}_{#1}^{({#2})}}}
        }
}

\DeclareDocumentCommand\ccc{ g g }{%
        \IfNoValueTF {#1} {\mathbf{q}} {
            \IfNoValueTF {#2} {{\mathbf{q}_{#1}}}{{\mathbf{q}_{#1}^{(#2)}}{}}
        }
}

\DeclareDocumentCommand\constc{ g }{%
        \IfNoValueTF {#1} {c} {
            {c^{({#1})}}
        }
}

\DeclareDocumentCommand\setz{ g g }{%
        \IfNoValueTF {#1} {\mathcal{Z}} {
            \IfNoValueTF {#2} {\mathcal{Z}^{(#1)}}{\mathcal{Z}^{(#1)}_{#2}}
        }
}

\DeclareDocumentCommand\setzp{ g g }{%
        \IfNoValueTF {#1} {\mathcal{Z^+}} {
            \IfNoValueTF {#2} {\mathcal{Z}^{+(#1)}}{\mathcal{Z}^{+(#1)}_{#2}}
        }
}

\DeclareDocumentCommand\setzn{ g g }{%
        \IfNoValueTF {#1} {\mathcal{Z^-}} {
            \IfNoValueTF {#2} {\mathcal{Z}^{-(#1)}}{\mathcal{Z}^{-(#1)}_{#2}}
        }
}

\DeclareDocumentCommand\tsetz{ g g }{%
        \IfNoValueTF {#1} {\tilde{\mathcal{Z}}} {
            \IfNoValueTF {#2} {\tilde{\mathcal{Z}}^{(#1)}}{\tilde{\mathcal{Z}}^{(#1)}_{#2}}
        }
}

\DeclareDocumentCommand\tz{ g g }{%
        \IfNoValueTF {#1} {\tilde{z}} {
            \IfNoValueTF {#2} {\tilde{z}^{(#1)}}{\tilde{z}^{(#1)}_{#2}}
        }
}

\DeclareDocumentCommand\f{ g g }{%
        \IfNoValueTF {#1} {f} {
            \IfNoValueTF {#2} {f^{(#1)}}{f^{(#1)}_{#2}}
        }
}

\DeclareDocumentCommand\lf{ g g }{%
        \IfNoValueTF {#1} {\underline{f}} {
            \IfNoValueTF {#2} {\underline{f}^{(#1)}}{\underline{f}^{(#1)}_{#2}}
        }
}

\def\eqref#1{Eq.~(\ref{#1})}

\def\1{\bm{1}}

\DeclareMathAlphabet{\mathsfit}{\encodingdefault}{\sfdefault}{m}{sl}
\SetMathAlphabet{\mathsfit}{bold}{\encodingdefault}{\sfdefault}{bx}{n}

\def\gC{{\mathcal{C}}}

\def\gF{{\mathcal{F}}}
\def\gG{{\mathcal{G}}}

\def\gL{{\mathcal{L}}}

\def\gS{{\mathcal{S}}}

\DeclareMathOperator*{\argmax}{arg\,max}

\usepackage{adjustbox}
\usepackage{booktabs}
\usepackage{multibib}
\usepackage{float}
\usepackage[ruled]{algorithm2e}
\usepackage{graphicx}
\usepackage{booktabs}
\usepackage{multirow}
\usepackage{multicol}
\usepackage[dvipsnames]{xcolor}
\usepackage[symbol]{footmisc}

\title{Improve Video Representation with Temporal Adversarial Augmentation}

\author{
Jinhao Duan$^1$
\and
Quanfu Fan$^2$\thanks{Part of work done while the author worked at MIT-IBM Watson AI lab. $\,^\dagger$Equal corresponding author.} \and
Hao Cheng$^{3}$\and
Xiaoshuang Shi$^{4\dagger}$ \And
Kaidi Xu$^{1\dagger}$
\affiliations
$^1$Drexel University \\
$^2$Amazon\\
$^3$The Hong Kong University of Science and Technology (Guangzhou)\\
$^4$University of Electronic Science and Technology of China \\
\emails
jd3734@drexel.edu,
quanfu@amazon.com,
hcheng046@connect.hkust-gz.edu.cn,
xsshi2013@gmail.com,
kx46@drexel.edu
}

\begin{document}

\maketitle

\begin{abstract}
Recent works reveal that adversarial augmentation benefits the generalization of neural networks (NNs) if used in an appropriate manner.
In this paper, we introduce \textbf{T}emporal Adversarial \textbf{A}ugmentation (TA), a novel video augmentation technique that utilizes temporal attention.
Unlike conventional adversarial augmentation, TA is specifically designed to shift the attention distributions of neural networks with respect to video clips by maximizing a temporal-related loss function.
We demonstrate that TA will obtain diverse temporal views, which significantly affect the focus of neural networks. Training with these examples remedies the flaw of unbalanced temporal information perception and enhances the ability to defend against temporal shifts, ultimately leading to better generalization. 
To leverage TA, we propose \textbf{T}emporal Video \textbf{A}dversarial \textbf{F}ine-tuning (TAF) framework for improving video representations.
TAF is a model-agnostic, generic, and interpretability-friendly training strategy. 
We evaluate TAF with four powerful models (TSM, GST, TAM, and TPN) over three challenging temporal-related benchmarks (Something-something V1\&V2 and diving48).
Experimental results demonstrate that TAF effectively improves the test accuracy of these models with notable margins
without introducing additional parameters or computational costs. As a byproduct, TAF also improves the robustness under out-of-distribution (OOD) settings. Code is available at \url{https://github.com/jinhaoduan/TAF}.
\end{abstract}

\section{Introduction}


Deep learning has achieved significant successes in multiple domains~\cite{yuan2020attribute,shi2020loss,yuan2023dde,cao2023comprehensive}. 
However, adversarial examples have been widely recognized as a serious threat to neural networks (NNs)~\cite{xu2018structured,xu2020adversarial}.
Imperceptible distortions created by advanced adversarial attack algorithms can easily manipulate the decision of well-trained neural networks.
This issue would be more critical for security-sensitive scenarios, such as biological identification~\cite{dong2019efficient} and autonomous~\cite{wang2022poster}.
However, recent works also reveal that adversarial examples could benefit NNs, if used in the appropriate manner.
For instance, adversarial examples could be the special cases when perceiving the category decision boundaries~\cite{tanay2016boundary}.
Also, by regarding adversarial examples as special augmentations, jointly utilizing adversarial examples and natural examples during the training will ameliorate the generalization of NNs~\cite{xie2020adversarial,chen2021robust}.

\begin{figure*}[ht]
\centering
\includegraphics[width=0.9\linewidth]{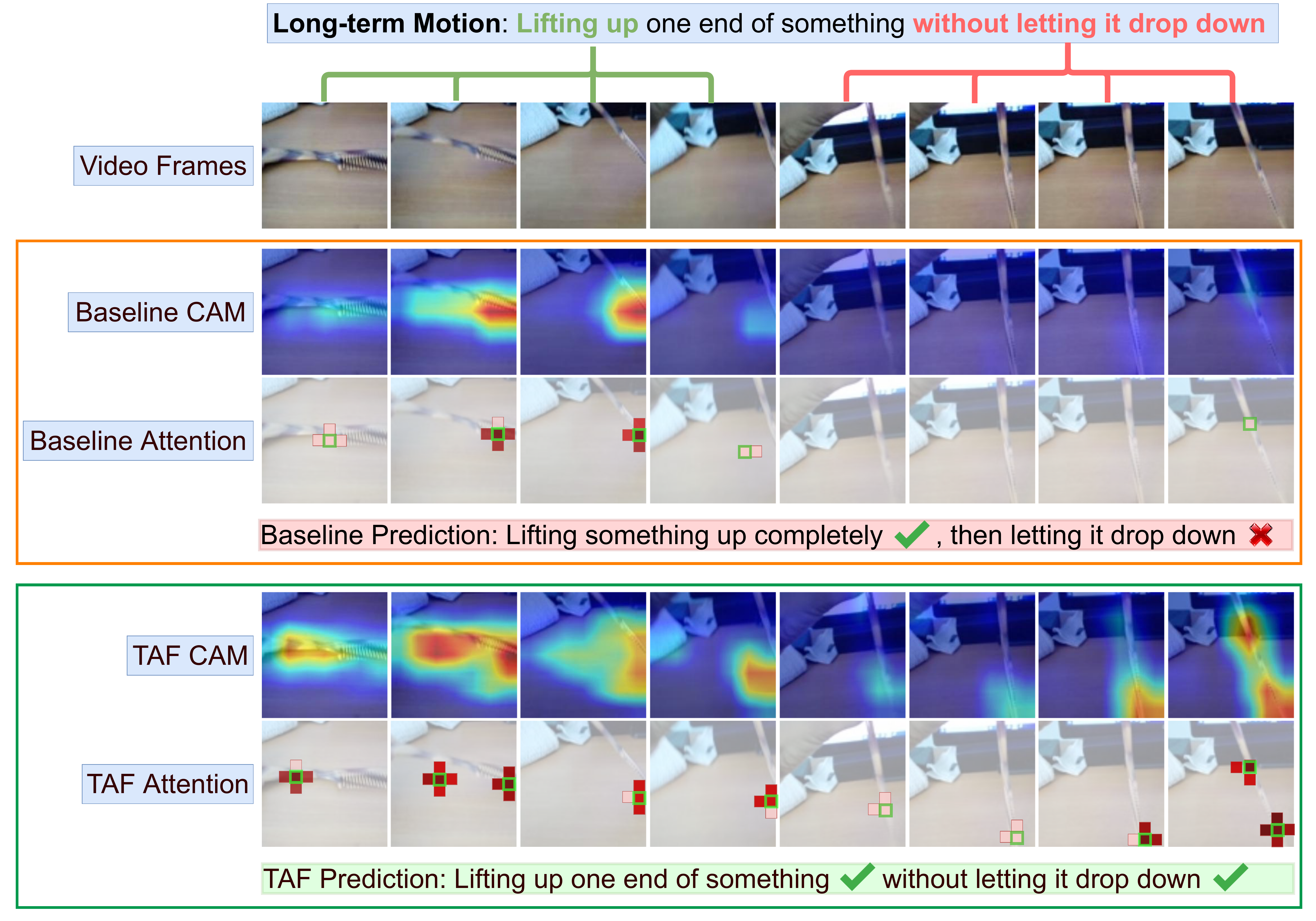}
\caption{
TAF enables models to capture global temporal cues by broadening and balancing the temporal attention distributions. As a result, the model is able to capture both the ``lifting'' and ``holding'' parts of the video, whereas the baseline model without TAF can only focus on the "lifting" process and ignores the "holding" parts.
}
\label{fig:motivation}
\end{figure*}

Temporal modeling is the decisive procedure for video understanding tasks~\cite{wang2016temporal}.
Recently, various modules are proposed to capture temporal information.
For example, equipping networks with temporal convolution operations~\cite{lin2019tsm,luo2019grouped,carreira2017quo} and local/global attention mechanisms~\cite{wang2018non,fan2019blvnet} are the most common practices.
Although these methods make great progress on this issue, the main concern is that such strategies tend to achieve narrow and overly centered temporal attention.
No mechanisms guarantee the surrounding temporal clues, which may contain valuable information, will also be fully explored equivalently.
Figure~\ref{fig:motivation} shows that current state-of-the-art model inferences only rely on a few frames, which ignores the following ``\textbf{holding}'' part and provides the wrong prediction. 
This property helps to converge training samples but might result in temporally over-fitting and hurts the generalization. 
Therefore, we may ask: do balanced temporal attention distributions help with the generalization?

Training with adversarial augmentation is a promising scheme to adaptively regularize the temporal attention distributions of NNs.
On the one hand, considering that adversarially augmented examples share the same semantic contents as natural examples, training with these examples will keep the consistency and stability of learning targets.
On the other hand, the constructed perturbations can largely affect the behavior of NNs, which provides an opportunity to concretely rectify models according to the needs.
Motivated by that, we propose \textbf{T}emporal Adversarial \textbf{A}ugmentation (TA) to address the unbalanced perception of temporal information. 
Different from conventional adversarial augmentation, TA is specifically designed to perturb the temporal attention distributions of NNs. 
Concretely, TA utilizes Class Activation Mapping (CAM)-based temporal loss function to represent the temporal attention distributions w.r.t. video clips, and disturb NNs' temporal views by maximizing the temporal loss function. 
In this way, videos augmented by TA will obtain diverse temporal attention (Figure~\ref{fig:adversarial examples}). Training with temporally augmented examples will remedy the defect of unbalanced attention assignation. Our contributions can be summarized as the following:

\begin{itemize}
    \item We introduce \textbf{T}emporal Adversarial \textbf{A}ugmentation (TA). TA changes the temporal distributions of video clips, and provides more temporal views for video understanding models.
    \item We propose \textbf{T}emporal Video \textbf{A}dversarial \textbf{F}ine-tuning (TAF) framework to regularize the attention distribution of networks by utilizing temporal adversarial augmentation. TAF is a model-agnostic, generic, and interpretability-friendly training scheme. This is the first work to improve video understanding models by explicitly utilizing adversarial machine learning.
    \item TAF is performed on four powerful video understanding models, including TSM~\cite{lin2019tsm}, TAM~\cite{fan2019blvnet}, GST~\cite{luo2019grouped}, and TPN~\cite{yang2020temporal}, and evaluated on three temporal related benchmarks (Somthing-something V1 \& V2 and Diving48). Experimental results demonstrate that TAF can significantly boost test accuracy, without any additional parameters and computational costs.
    \item TAF is evaluated under the out-of-distribution~\cite{hendrycks2018benchmarking} settings and it effectively boosts robustness in defending naturally corrupted data with notable margins.
    
\end{itemize}

\begin{figure*}[ht]
\centering
\includegraphics[width=0.8\linewidth]{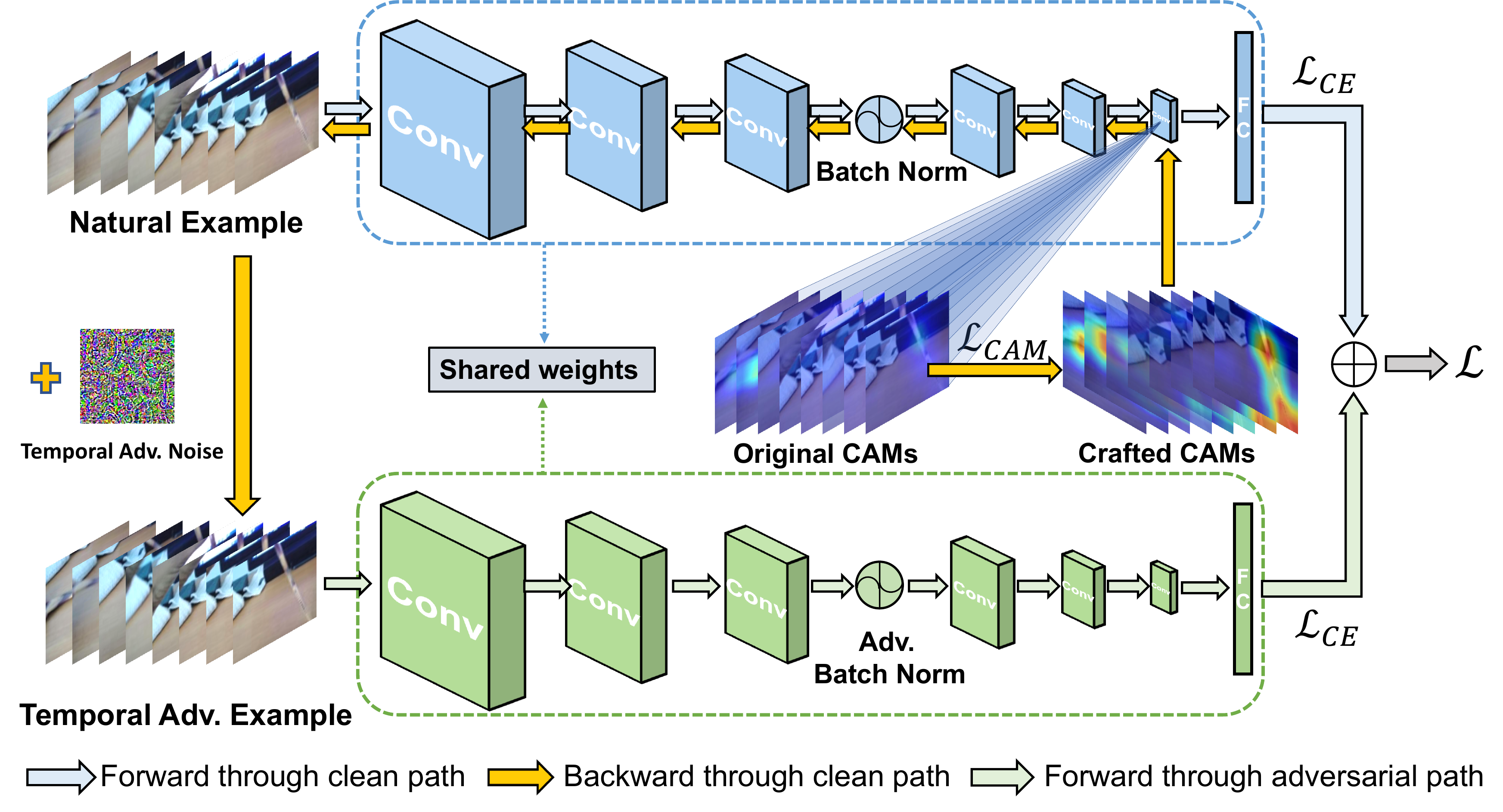}
\caption{The overall pipeline of TAF. The upper network (clean path) and the lower network (adversarial path) share weights except for batch normalization layers.}
\label{fig:architecture}
\end{figure*}


\section{Related Works}

\subsection{Adversarial Machine Learning for Good.}
Enhancing NNs by taking advantage of adversarial machine learning is being popular.
One of the main contributions in this field is revealing the relationship between robustness and generalization~\cite{tsipras2018robustness,su2018robustness}.
\cite{tsipras2018robustness} theoretically proves that there is an inherent gap between robustness and generalization.
Similar conclusions are proposed in~\cite{su2018robustness} with empirical evidence.
In applications, \cite{xie2020adversarial,chen2021robust} prove that by utilizing auxiliary batch normalization layers, adversarial examples could benefit the generalization and improve image classification tasks.
Besides, adversarial examples show great potential in understanding and interpreting NNs~\cite{ignatiev2019relating,boopathy2020proper}.
\cite{ignatiev2019relating} reveals that adversarial examples and explanations can be connected by the general formal of hitting set duality.
\cite{boopathy2020proper} propose interpretability-aware defensive mechanism to achieve both robust classification and robust interpretation.

\subsection{Robustness of Video Understanding.}
Video understanding, as one of the most important vision tasks, has been promoted to a new level with the help of NNs. ~\cite{carreira2017quo,xie2018rethinking,lin2019tsm,wang2016temporal} propose to improve video representations with NNs constructed by different architectures (e.g., 2D/3D structures, recurrent architectures). ~\cite{arnab2021vivit,wei2022masked} prove that Transformer architectures benefit video understanding tasks a lot. Besides, boosting temporal modeling by incorporating NNs with additional attention mechanisms is also practical and popular~\cite{wang2018non,fan2019blvnet}.
In terms of robustness, various attack and defense algorithms are proposed in video scenarios. ~\cite{li2018adversarial,chen2021appending} show that video understanding models can also be easily manipulated by imperceptible perturbations. Furthermore, ~\cite{li2021adversarial} proves that video models can be attacked under different settings (e.g., black-box attack and transfer attack). On the other side, common defense techniques (e.g., adversarial training~\cite{kinfu2022analysis}, adversarial detection~\cite{thakur2022pat}) are also compatible with video understanding models.

\subsection{Class Activation Mapping.}
Class Activation Mapping (CAM)~\cite{zhou2016learning} is one of the most popular approaches in understanding and interpreting the attention distributions of NNs.
CAM measures the importance of channels by global average pooling and weighted fusion.
After that, a series of gradient-based~\cite{selvaraju2017grad,chattopadhay2018grad} and gradient-free~\cite{petsiuk2021black} CAM solvers are also introduced.
Gradient-based CAM uses either first-order~\cite{selvaraju2017grad} or second-order~\cite{chattopadhay2018grad} gradient information to perceive the interested regions of NNs. 
Considering GradCAM~\cite{selvaraju2017grad} is compatible with various model structures, and it is more effective than gradient-free CAM, we choose it as the CAM producer in the rest of this paper.

\section{Approach}
In this section, we describe how to capture balanced temporal information with the proposed Temporal Video Adversarial Fine-tuning (TAF) framework.
Firstly, we revisit the vanilla adversarial augmentation.
Then, detailed illustrations of TA along with the CAM-based metrics are provided.
At last, we outline the training and testing protocols of the TAF framework.
The pipeline of TAF is shown in Figure~\ref{fig:architecture}.

\subsection{Vanilla Adversarial Augmentation}\label{sec:vanilla adv augmentation}
Adversarial augmentation derives from adversarial perturbation, a kind of imperceptible noise that can easily disturb the predictions of well-trained NNs. ~\cite{xie2020adversarial,chen2021robust} show that adversarial perturbations can be seen as special augmentations to improve generalization and robustness in image-based vision tasks.

For a given model, $\gF$, parameterized by weights $\theta$ and input $X \in \mathbb{R}^{C \times H \times W}$ with $C$ channels and resolution $H \times W$, the adversarially augmented example, $X ^\prime$, is defined as:
\begin{equation}
\begin{aligned}
    X^\prime = X + \delta = X + \argmax_{\delta \in [-\epsilon, \epsilon]} {\gL(X + \delta, y)},
\end{aligned}
\end{equation}
where $\delta$ is the adversarial noise solved by either single-step (e.g., FGSM~\cite{goodfellow2014explaining}) or iterative (e.g., PGD~\cite{kurakin2016adversarial}) attack algorithms, $\epsilon$ is the attack budget, and $\gL$ is the conventional classification loss (i.e., Cross-entropy Loss).

Vanilla adversarial augmentation is an effective technique for image tasks, but it is not well-suited for video scenarios. Video understanding models often suffer from serious overfitting issues, with over 40\% overfitting gaps (i.e., top-1 training accuracy vs top-1 validation accuracy) observed on the Something-something V1 dataset. This severe overfitting suggests that a lot of generalization-irrelevant noise is introduced during the training. For neural networks, the loss function plays a crucial role in determining which features or information are absorbed. Therefore, using a classification loss alone may propagate these irrelevant noises back to the adversarial perturbation, which ultimately harms the generalization of neural networks. 

\subsection{Temporal Adversarial Augmentation}\label{sec:temporal adversarial examples}


To address this issue, the proposed Temporal Adversarial Augmentation (TA) utilizes a CAM-based temporal loss function to leverage temporal attention-related information solely, which is one of the most fundamental and essential features of videos. Here we show how to incorporate temporal information into adversarial augmentation.

For a given model, $\gF$, parameterized by weights $\theta$ and video volume $X \in \mathbb{R}^{T \times N_c \times H \times W}$ with $T$ frames, $N_c$ channels and $H\times W$ resolution, we first consider the CAMs of model $\gF$ w.r.t. frame $X_i$,
\begin{equation}
\begin{aligned}
    X_i^{\gC} = \gG^{\gC}(\theta, X_i, \hat{y}) &= \frac{1}{HW} \sum^{H} \sum^{W} \frac{\partial \gL_{ce}(\theta, X, \hat{y})}{\partial \gF^{'}(X_i)}\gF^{'}(X_i), \\ & i=1, \dots, T,
\end{aligned}
\end{equation}
where $\gF^{'}$ refers to the subnetwork of $\gF$ from the input layer to the final convolutional layer and $\gL_{ce}(\cdot,\cdot,\cdot)$ is Cross-entropy loss.
$X^{\gC}_i$ represents the CAM of the $i$-th frame.
To discriminate the importance of each frame, we normalize each CAM and define the overall CAM on video $X$ as
\begin{equation}
\small
    \hat{X}_{i}^{\gC} = \frac{X_i^{\gC} - \min(X^{\gC})}{\max(X^{\gC})} \, \,\,\, \, i=1,\dots,T ,
\end{equation}
where $\min(\cdot)$ and $\max(\cdot)$ refer to the minimal and maximal values among all CAMs and $X^{\gC} = \{ 
X_{1}^{\gC}, \cdots, X_{T}^{\gC} \}$.

\begin{figure}[ht]
\centering
\includegraphics[width=1\columnwidth]{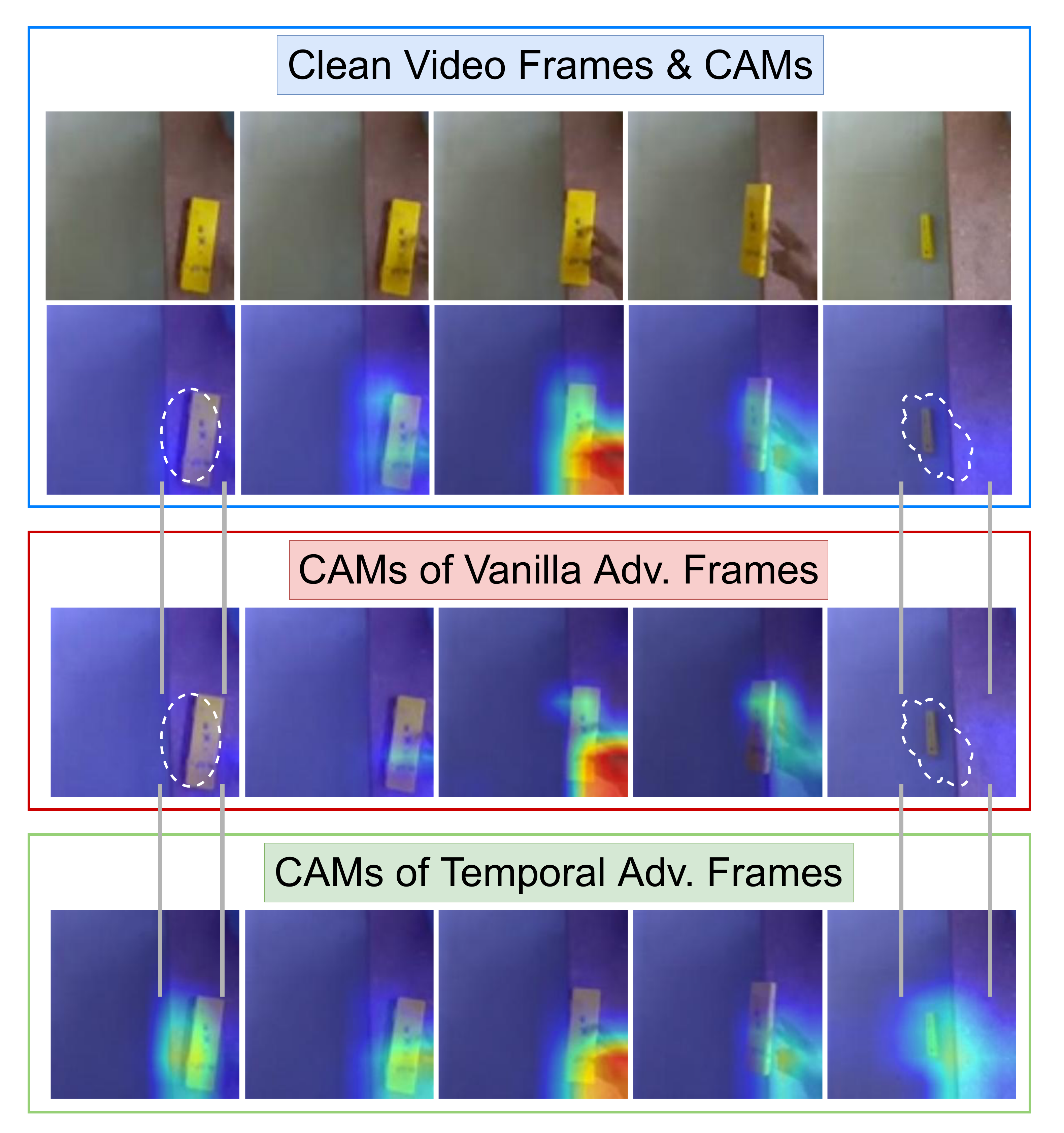}

\caption{Adversarially augmented examples and the corresponding CAMs.
Vanilla adversarial augmentation does not substantially affect the temporal attention distribution, while our temporal adversarial augmentation changes it significantly. This highlights the benefits of TAF in terms of broadening and balancing the temporal attention distributions.
}\label{fig:adversarial examples}
\end{figure}

We balance temporal attention distribution by amplifying those ``unimportant'' frames (i.e., frames with smaller CAM values).
Concretely, we sort $\hat{X}_{i}^{\gC}$ in ascending order according to the sum of CAM values at each frame:
\begin{equation}
    \Sigma {\hat{X}_{\pi 1}^{\gC}} < \Sigma {\hat{X}_{\pi 2}^{\gC}} < \cdots < \Sigma {\hat{X}_{\pi N}^{\gC}} < \cdots < \Sigma {\hat{X}_{\pi K}^{\gC}}
\end{equation}
 The top $N$ frames with the smallest CAM values are selected as the \textit{non-key frames}.
Then, the CAM-based loss is defined as
\begin{equation}
\small
    \gL_{\gC} = \frac{1}{N} \sum^{N}_{i} {\hat{X}_{\pi i}^{\gC}}.
\end{equation}
By maximizing $\gL_{\gC}$, global temporal attention will be reassigned to those unimportant frames, which balances the attention distribution.

To generate the final temporal augmentation, we utilize the popular iterative gradient sign method (i.e. PGD~\cite{kurakin2016adversarial}) and update $X$ in $K$ iteration steps with the formulation
\begin{equation}\label{eq:attack}
\begin{aligned}
    X_i^{k+1} = & \Pi_{X\pm \epsilon}(X_i^k + \beta * sgn(\nabla_{X_{i}^K} \gL_{\gC}(\theta, X_i, \hat{y}))), \\ 
    & i=1,\dots,T, \,\,\, k=1,\dots,K,
\end{aligned}
\end{equation}
where $\epsilon$ refers to the attack budget under $\ell_{\inf}$ constraint and $\Pi_{X \pm \epsilon}(\cdot)$ projects tensor back to $[X-\epsilon, X+\epsilon]$.
$\beta$ represents the attack step size.
In terms of the targeted label $\hat{y}$, we assign it based on the example's prediction.
For correctly classified samples, models are expected to be robust against temporal shifting.
Therefore, a random label is set to $\hat{y}$, which will generate diverse temporal views.
However, for incorrectly classified samples, these are hard examples for the model and we choose to reduce the difficulties of these examples by correctly boosting the temporal attention~\cite{lee2021anti}.
In this situation, we set $\hat{y}$ with the true label.

As shown in Figure~\ref{fig:adversarial examples}, by shifting temporal attention of models, temporally augmented video clips have diverse temporal views compared with conventional adversarial augmentation.
Training with these examples can be seen as a regularization applied to the NNs for encouraging diverse inference, which results in better generalization to unseen samples.

\subsection{Adversarial Fine-tuning Framework for Video Understanding}\label{sec:TAF}

TAF jointly utilizes both natural examples and adversarially augmented examples with the following optimization object:
\begin{equation}\label{eq:TAF learning target}
    \min_\theta \mathbb{E}_{(x, y) \sim D} \left[\alpha * \gL_{ce}(\theta, x, y) + (1-\alpha) * \gL_{ce}(\theta, x^{K}, y)\right],
\end{equation}
where $\gL_{ce}$ is the Cross-entropy loss function, and $x^{K}$ is the temporally augmented data based on Eqn.~\ref{eq:attack}. $\alpha$ is used to control the contribution of loss items.
Besides, inspired by~\cite{xie2020adversarial}, we adopt additional normalization layers (i.e., Batch Normalization~\cite{ioffe2015batch}) to deal with the distribution mismatching between adversarial examples and natural examples.
We name the forward path with original normalization layers and additional normalization layers as \textit{clean path} and \textit{adversarial path}, respectively.

For training, the natural examples will be first passed through the clean path and get two outputs: clean logits and CAM feature maps.
Then, temporally augmented examples are generated by performing Eq.~\ref{eq:attack} on CAM feature maps and back-propagating through the clean path to the input example.
Next, pass the augmented examples through the adversarial path and get adversarial logits.
Finally, optimizing both clean path and adversarial path with classification loss.
For inference, we drop adversarial normalization layers and get the final prediction with the clean path.
The pseudo-code of TAF is shown in Appendix~\ref{appendix:pseudo-code}.



\begin{table}[ht]
    \centering
    \resizebox{1.0\columnwidth}{!}{
    \begin{tabular}{c|l|cc|ll}
        \toprule
        Dataset & \multicolumn{1}{c|}{Method} & Backbone & frames & Top-1(\%) & Top-5(\%)  \\
        \midrule
        \multirow{14}{*}{SSV1} & TSM & resnet50 & 8 & 45.6 & 74.6 \\
        &TSM+TAF & resnet50 & 8      & \textbf{46.9(\textcolor{Green}{+1.3})} & \textbf{75.0(\textcolor{Green}{+0.4})}      \\
        &TSM     & resnet50 & 16     & 47.2          & 77.1      \\
        &TSM+TAF & resnet50 & 16     & \textbf{47.9(\textcolor{Green}{+0.7})} & \textbf{77.7(\textcolor{Green}{+0.6})}      \\
        \cmidrule(lr){2-6}
        &GST     & 3d-resnet50 & 8      & 46.6          & 75.6   \\
        &GST+TAF & 3d-resnet50 & 8      & \textbf{47.6(\textcolor{Green}{+1.0})} & \textbf{76.4(\textcolor{Green}{+0.8})}   \\
        &GST     & 3d-resnet50 & 16     & 48.4          & 77.2   \\
        &GST+TAF & 3d-resnet50 & 16     & \textbf{48.8(\textcolor{Green}{+0.4})} & \textbf{77.5(\textcolor{Green}{+0.3})}   \\
        &GST-L     & 3d-resnet50 & 8      & 47.0          & 76.1 \\
        &GST-L+TAF & 3d-resnet50 & 8      & \textbf{47.7(\textcolor{Green}{+0.7})} & \textbf{76.3(\textcolor{Green}{+0.2})} \\
        \cmidrule(lr){2-6}
        &TAM     & resnet50 & 8      & 46.2          & 75.4      \\
        &TAM+TAF & resnet50 & 8      & \textbf{46.6(\textcolor{Green}{+0.4})} & 75.4      \\
        \cmidrule(lr){2-6}
        &TPN     & resnet50 & 8      & 48.0          & 77.2      \\
        &TPN+TAF & resnet50 & 8      & \textbf{49.1(\textcolor{Green}{+1.1})} & \textbf{78.0(\textcolor{Green}{+0.8})}      \\
        \midrule
        \multirow{8}{*}{SSV2} & TSM & resnet50 & 8 & 58.9 & 85.5 \\
        &TSM+TAF & resnet50 & 8      & \textbf{59.8(\textcolor{Green}{+0.9})} & \textbf{86.0(\textcolor{Green}{+0.5})} \\
        &TSM     & resnet50 & 16     & 61.1          & 86.8  \\
        &TSM+TAF & resnet50 & 16     & \textbf{62.0(\textcolor{Green}{+0.9})} & \textbf{87.3(\textcolor{Green}{+0.5})} \\
        \cmidrule(lr){2-6}
        &GST     & 3d-resnet50 & 8      & 61.3          & 87.2         \\
        &GST+TAF & 3d-resnet50 & 8      & \textbf{61.7(\textcolor{Green}{+0.4})} & \textbf{87.4(\textcolor{Green}{+0.2})}  \\
        \cmidrule(lr){2-6}
        &TPN     & resnet50 & 8      & 61.6          & 87.7        \\
        &TPN+TAF & resnet50 & 8      & \textbf{62.1(\textcolor{Green}{+0.5})} & \textbf{88.3(\textcolor{Green}{+0.6})}   \\
        \midrule
        \multirow{8}{*}{D48} & TSM & resnet50 & 8 & 78.5 & 97.3 \\
        &TSM+TAF & resnet50      & 8      & \textbf{79.1(\textcolor{Green}{+0.6})} & \textbf{97.6(\textcolor{Green}{+0.3})} \\
        \cmidrule(lr){2-6}
        &GST     & 3d-resnet50   & 8      & 73.9          & 96.7 \\
        &GST+TAF & 3d-resnet50   & 8      & \textbf{74.7(\textcolor{Green}{+0.8})} & \textbf{96.9(\textcolor{Green}{+0.2})} \\
        \cmidrule(lr){2-6}
        &TAM     & resnet50      & 8      & 75.1          & 96.3  \\
        &TAM+TAF & resnet50      & 8      & \textbf{75.8(\textcolor{Green}{+0.7})} & \textbf{97.1(\textcolor{Green}{+0.8})} \\
        \cmidrule(lr){2-6}
        &TPN     & resnet50      & 8      & 80.2          & \textbf{98.4}  \\
        &TPN+TAF & resnet50      & 8      & \textbf{80.9(\textcolor{Green}{+0.7})} & 98.0  \\
        \bottomrule
    \end{tabular}
    }
    \vspace{-2mm}
    \caption{Evaluations on Something-something V1 and V2 (SSV1 and SSV2), Diving48 (D48) benchmarks.}
    \label{tab:main_result}
\end{table}

\section{Experiments}
In this section, we investigate the effectiveness of TAF through comprehensive experiments.
Concretely, we first introduce our experiment settings, including datasets, baselines, and implementation details.
The evaluations on multiply state-of-the-art models and challenging benchmarks are followed.
Then, a series of ablation studies are conducted, including comparing with vanilla adversarial augmentation, impacts of $\alpha$ and attacking settings, relieving overfitting, and the discussion about computational costs.
Moreover, we qualitatively analyze TAF by providing representative visualizations.
At last, we examine the models' ability against naturally corrupted data~\cite{hendrycks2018benchmarking}, i.e., out-of-distribution (OOD) robustness.




\subsection{Datasets and Baselines}
We evaluate TAF on three popular temporal datasets: Something-something V1\&V2~\cite{goyal2017something}, Diving48~\cite{li2018resound}. 
Something-something V1\&V2 are large-scale challenging video understanding benchmarks consisting of 174 classes. Diving48 is a fine-grained temporal action recognition
dataset with 48 dive classes.
The detailed introduction of these datasets can be found in Appendix~\ref{appendix:dataset_intro}

The reason why we choose these three datasets is that TAF aims to tackle temporal modeling issues. Both Something-something and Diving48 are the most challenging benchmarks in this field~\cite{lin2019tsm,jiang2019stm}.
Only action descriptions are reserved in these datasets, without introducing scene-related knowledge that enforces the model to learn the temporal information.
In this way, TAF can be fairly and entirely evaluated.

{\bf Baselines.} To evaluate the effectiveness of TAF, we conduct experiments on four powerful action recognition models: TSM~\cite{lin2019tsm}, GST~\cite{luo2019grouped}, TAM~\cite{fan2019blvnet} and TPN~\cite{yang2020temporal}.
These models cover the most representative action recognition methodologies: 2DConvNets-based recognition~\cite{lin2019tsm,fan2019blvnet,yang2020temporal}, 3DConvNet-based recognition~\cite{luo2019grouped} and attention-based recognition~\cite{fan2019blvnet,yang2020temporal}.
We train all the baseline models by strictly following the training protocols provided by their official codebases.

{\bf Fine-tuning.} For fine-tuning, we load pre-trained weights and keep training 15 epochs with TAF.
We conduct 3 trials for each experiment and report the mean results.
The initial training settings (e.g., learning rate, batchsize, dropout, etc.) are the same as the status when the pre-trained models are logged.
The learning rates are decayed by a factor of 10 after 10 epochs.
We set $\alpha$ as 0.7, and the number of attacked frames $N$ as 8 or 16 according to the input temporal length.
Note that considering most of the baseline models did not conduct experiments on Diving48, we adopt the same training settings as on Something-something datasets.

{\bf Inference.} For fairness and convenience, all the performances reported in this paper are evaluated on 1 center crop and 1 clip, with input resolution 224 $\times$ 224.

\begin{table*}
\centering
\begin{subtable}{0.45\linewidth}

\resizebox{\linewidth}{!}{
    \begin{tabular}{c|cccccc}\toprule
        Method & Scratch & Fine-t. & CELoss & CAMLoss & Top-1 & $\Delta$ \\ \midrule
        baseline & $\surd$ & &  &  & 45.6 & - \\
        AdvProp~[51] &  $\surd$ & & $\surd$ & & 44.5 & -1.1 \\
        CE-based & &$\surd$ & $\surd$ &  & 46.1 &  +0.5\\ 
        \midrule 
        TAF & & $\surd$ & & $\surd$ & \textbf{46.9} & +1.3  \\ 
        \bottomrule
    \end{tabular}
}
\caption{Scratch and Fine-t. stand for training from scratch and fine-tuning from pretrained models. CELoss/CAMLoss refers to the objective function used to generate adversarial perturbations.}
\label{tab:ce v.s. cam}
\end{subtable}
\hfil
\begin{subtable}{0.23\linewidth}

\resizebox{\linewidth}{!}{
    
    \begin{tabular}{c|cc}\toprule 
		Method                &   Top-1 (\%)          & Top-5 (\%) \\ \midrule
		baseline        & 45.6                  & 74.6      \\
  \midrule
		$\alpha$ = 0.2  &     46.3              &     75.1      \\
		$\alpha$ = 0.5  &     46.7              &     75.1      \\
		$\alpha$ = 0.7  &     \textbf{46.9}     &     75.0      \\
		$\alpha$ = 0.8  &     46.5              &     74.8      \\ \bottomrule
	\end{tabular}
}
\caption{Impacts of applying different $\alpha$ when optimizing objectives.}\label{tab:alpha}
\end{subtable}
\hfil
\begin{subtable}{0.23\linewidth}

\resizebox{\linewidth}{!}{
    \begin{tabular}{c|cc}\toprule
		Method                &   Top-1 (\%)          & Top-5 (\%) \\ \midrule
		baseline        & 45.6                  & 74.6      \\ 
        \midrule
		$\epsilon$ = 6 $K$ = 1  &     46.6              &     75.1      \\
		$\epsilon$ = 6 $K$ = 3  &     46.5              &     74.6      \\
		$\epsilon$ = 64 $K$ = 1  &     \textbf{46.9}     &     75.0      \\
		$\epsilon$ = 64 $K$ = 3  &     46.8              &     75.0      \\ \bottomrule
		\end{tabular}
}
\caption{Impacts of attacking settings. $\epsilon$ is scaled by 255. $K$ refers to the number of attack steps.}
\label{tab:attack settings}
\end{subtable}
\caption{Ablation study on Something-something V1 benchmark.}
\end{table*}

\setlength{\tabcolsep}{1.4pt}
\subsection{Comparisons with State-of-the-Art Models}\label{sec:sota}
The performances of TAF on Something-something V1 \& V2, and Diving48 are summarized in Table~\ref{tab:main_result}.
It is shown that TAF effectively improves on TSM, TAM, TPN, GST with 1.3\%, 0.4\%, 1.1\%, 1.0\% top-1 accuracy on Something-something V1, respectively. Similar promotions are also observed on the V2 dataset.
As for longer temporal inputs, we evaluate TAF in 16-frames settings and show that TAF augments TSM with 0.7\% and 0.9\% top-1 accuracy on V1 and V2, separately. The experiments of longer inputs on GST are also promising.
For Diving48, TAF improves TSM by 0.6\%, GST by 0.8\%, TAM by 0.7\% and TPN by 0.7\%.

\begin{figure}
\centering
\includegraphics[width=1\linewidth]{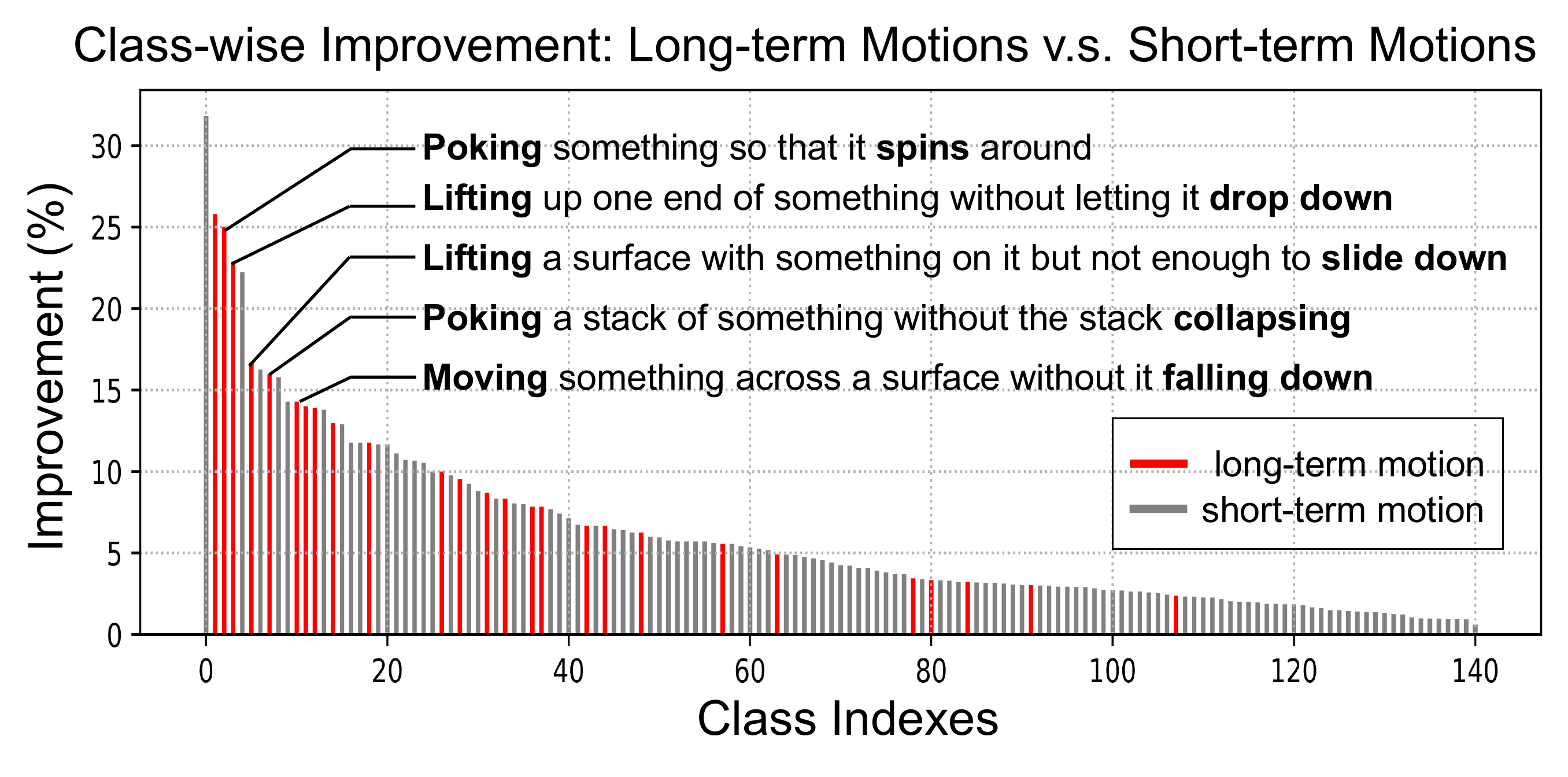}
\caption{
Class-wise improvements on Something-something V1 dataset.
TAF benefits more on the long-term motions.
}
\label{fig:long-term motions}
\end{figure}

\begin{figure}
\centering
  \includegraphics[width=1\linewidth]{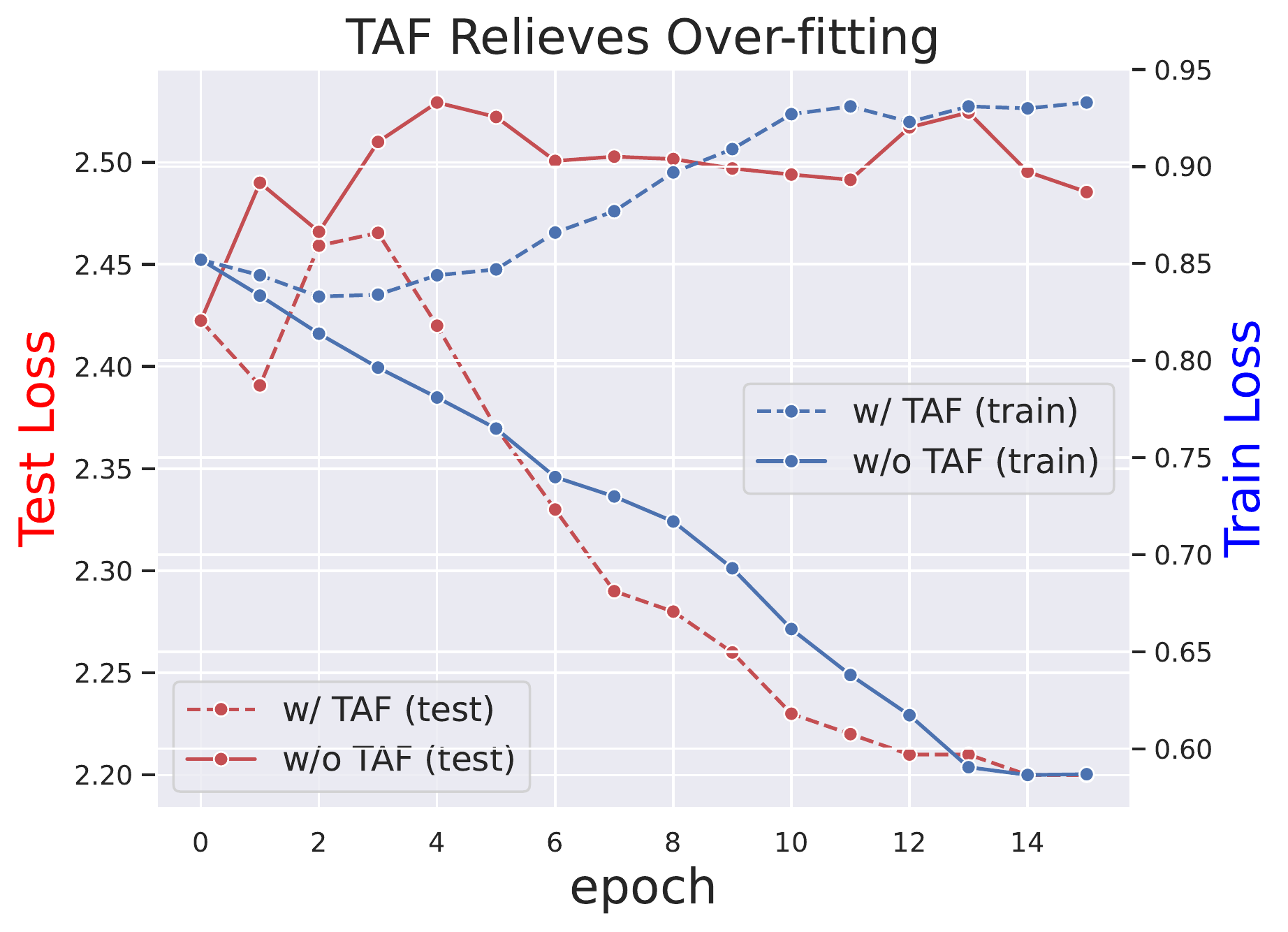}
\caption{Training and testing loss curves of TSM w/ and w/o TAF on Something-something V1. The testing loss is decreased while the training loss is increased when TAF is utilized, which shows that TAF effectively relieves over-fitting.}
\label{fig:curves}
\end{figure}

To further investigate the impact of TAF, class-wise improvements v.s. class indexes are summarized in Figure~\ref{fig:long-term motions}.
Since TAF benefits models by broadening the temporal attention distributions, we focus on the gains of motions requiring longer temporal knowledge, referred to as \textit{long-term motions}.
We define the long-term motions as actions consisting of at least two motions or phenomenons, such as ``\textbf{Lifting up} one end of something without letting it \textbf{drop down}''.
Based on our definition, 35 categories are recognized as the long-term motions out of 174 classes from Something-something benchmarks.
Appendix~\ref{appendix:long term motions} provides the complete list of long-term motions. 
The aggregation of red bars demonstrates that TAF substantially enhances the ability to capture long-term temporal cues: Over 10 long-term motions get improved largely (i.e., $\ge$ 10\%).

\subsection{Ablation Study}\label{sec:ablation}
We conduct comprehensive experiments exploring the impacts of TAF.
At first, we compare the proposed TAF with Cross-entropy based (CE-based) training strategy.
Then, we study the effects of $\alpha$ and the performances of TAF under different attacking settings (e.g., epsilon $\epsilon$, attack steps $T$).
The training loss and testing loss are provided to justify the regularization of overfitting.
Note that all ablation studies are performed on resnet50-TSM, with Something-something V1 as the dataset.
Appendix~\ref{appendix:ablation study} provides additional ablation studies and comparisons with other popular data augmentation methods.

\begin{table*}
\setlength{\tabcolsep}{4pt}
    \centering
    \resizebox{1.7\columnwidth}{!}{
    \begin{tabular}{l|ccc|ccc|cc}\toprule
              &  \multicolumn{3}{c|}{Noise} & \multicolumn{3}{c|}{Blur} & \multicolumn{2}{c}{Weather}   \\ 
              \cmidrule(lr){2-4}
              \cmidrule(lr){5-7}
              \cmidrule(lr){8-9}
		Model &  Gauss. & Impulse & Speckle & Gauss. & Defocus & Zoom & Snow & Bright \\ \midrule
		TSM   & 19.9/41.5 & 16.7/35.7 & 20.1/41.4 & 20.0/41.5 & 19.3/40.8 & 18.9/40.3 & 11.0/25.3 & 17.2/36.2 \\ 
		TSM+TAF  & \textbf{21.0}/\textbf{43.6} & \textbf{17.9}/\textbf{37.7} & \textbf{21.1}/\textbf{44.0} & \textbf{20.8}/\textbf{43.6} & \textbf{20.4}/\textbf{42.9} & \textbf{20.0}/\textbf{42.3} & \textbf{11.9}/\textbf{27.2} & \textbf{18.2}/\textbf{38.5}  \\ 
		TPN  & 24.7/48.8 & 17.3/37.6 & 24.4/48.8 & 25.3/49.6 & 24.7/48.7 & 24.3/48.2 & 13.6/31.0 & 19.5/40.1   \\ 
		TPN+TAF & \textbf{25.7}/\textbf{50.1} & \textbf{18.7}/\textbf{39.8} & \textbf{25.6}/\textbf{49.8} & \textbf{26.2}/\textbf{50.7} & \textbf{25.7}/\textbf{50.1} & \textbf{25.3}/\textbf{49.4} & \textbf{15.1}/\textbf{32.6} & \textbf{20.8}/\textbf{41.5}  \\ \bottomrule
		
		\end{tabular}
		}
    \caption{Evaluations of defending natural corruption. Performances are reported as Top-1(\%)/Top-5(\%).}
    \label{tab:natural corruption}
\end{table*}

\subsubsection{Comparison with Vanilla Adversarial Augmentation}
Results are placed in Table~\ref{tab:ce v.s. cam}.
It is shown that although CE-based method achieves a certain improvement (0.5\%) over baseline model, still TAF remarkably outperforms CE-based method (0.8\%) and baseline model (1.3\%).
As we mentioned in Sec.~\ref{sec:vanilla adv augmentation}, directly taking advantage of conventional classification loss will introduce lots of irrelevant noises into the adversarial perturbations, especially in heavy-overfitting situations.
Training with these examples benefits less on generalization.
However, TAF utilizes CAM-based loss function to filter all noise except temporal modeling knowledge.
Therefore, training with temporally augmented examples is more effective and more suitable for video understanding tasks.

\subsubsection{Impacts of $\alpha$ and attack settings}
Impacts of $\alpha$ and different attack settings are shown in Table~\ref{tab:alpha} and Table~\ref{tab:attack settings}.
It shows that incorporating with a proper portion of temporally augmented examples can effectively boost TSM, and the best performance is achieved when $\alpha$ is set to 0.7.
In terms of the impacts of attack settings, we investigate it by applying two schemes: small perturbation (i.e., $\epsilon = 6/255$, $\beta = 2/255 $) and large perturbation (i.e., $\epsilon = 64/255$, $\beta = 32/255 $) in Table~\ref{tab:attack settings}, along with single (i.e., $K = 1$) or multiply (i.e., $K = 3$) steps.
Generally, larger $\epsilon$ allows inject more temporal perturbations into natural examples and further achieves better temporal robustness.


\begin{figure}
    \centering
    
    \begin{subfigure}{.08\textwidth}
    \centering
    \includegraphics[width=\linewidth]{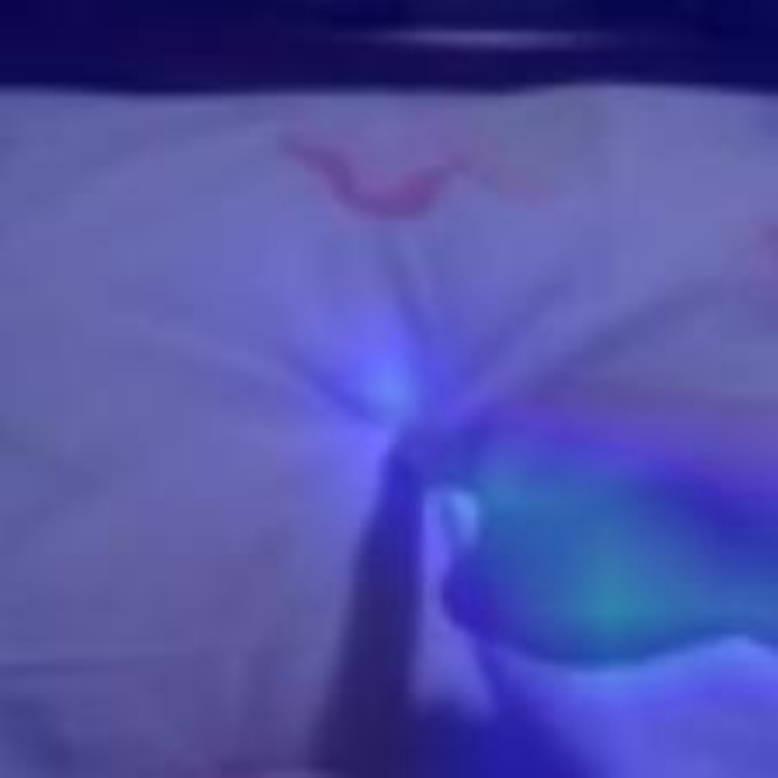}
    \end{subfigure}%
    \begin{subfigure}{.08\textwidth}
    \centering
    \includegraphics[width=\linewidth]{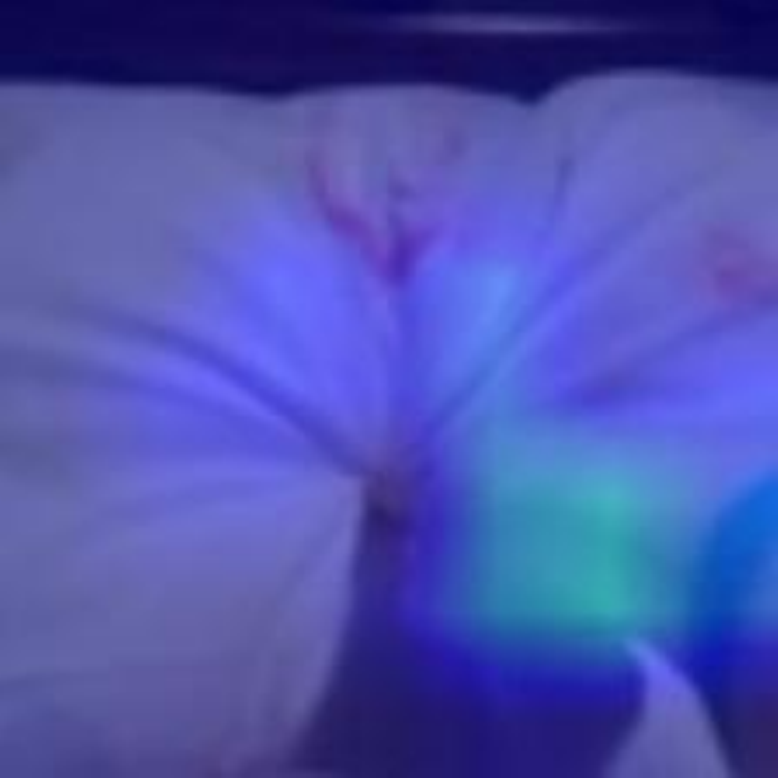}
    \end{subfigure}%
    \begin{subfigure}{.08\textwidth}
    \centering
    \includegraphics[width=\linewidth]{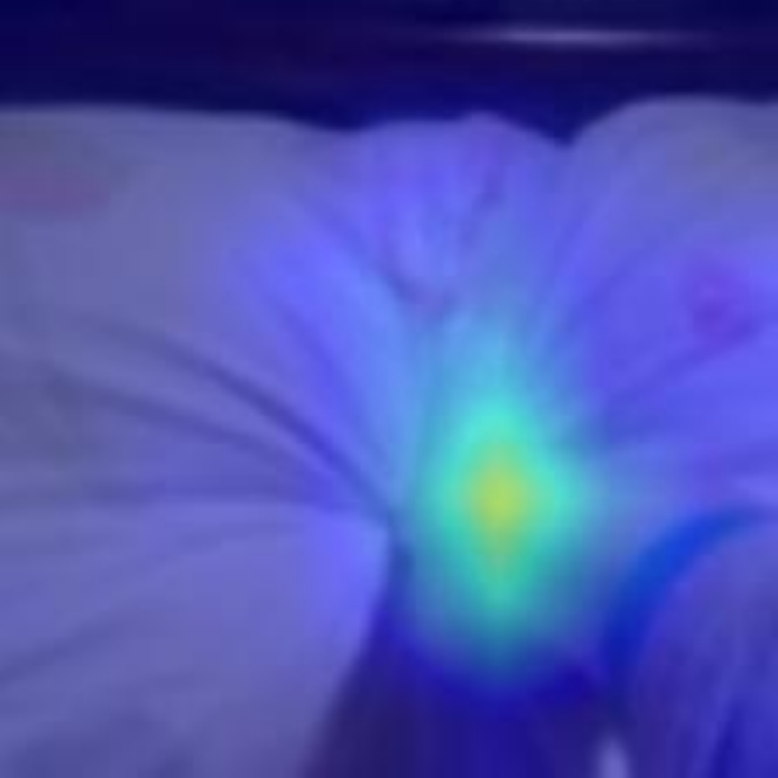}
    \end{subfigure}%
    \begin{subfigure}{.08\textwidth}
    \centering
    \includegraphics[width=\linewidth]{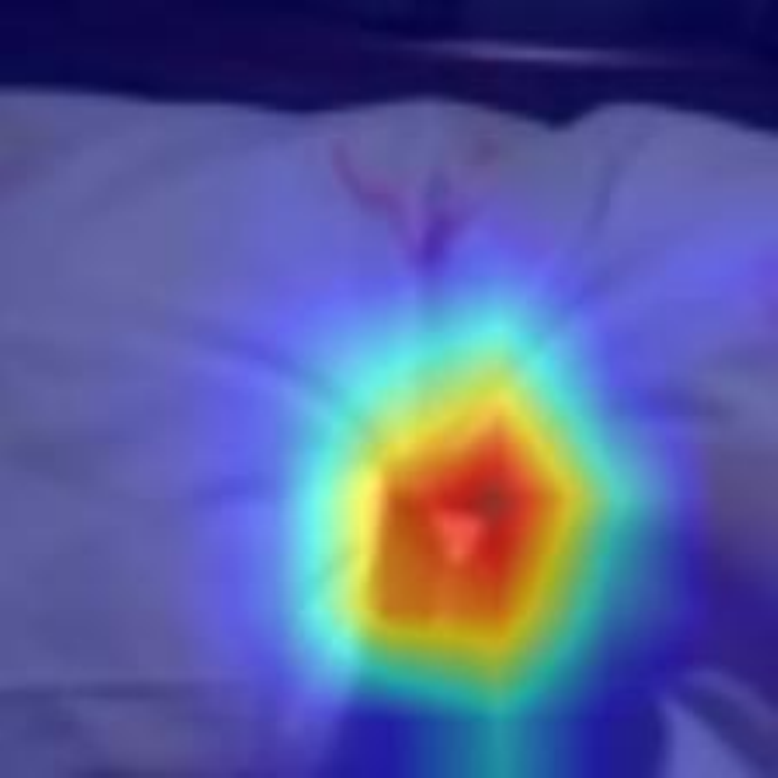}
    \end{subfigure}%
    \begin{subfigure}{.08\textwidth}
    \centering
    \includegraphics[width=\linewidth]{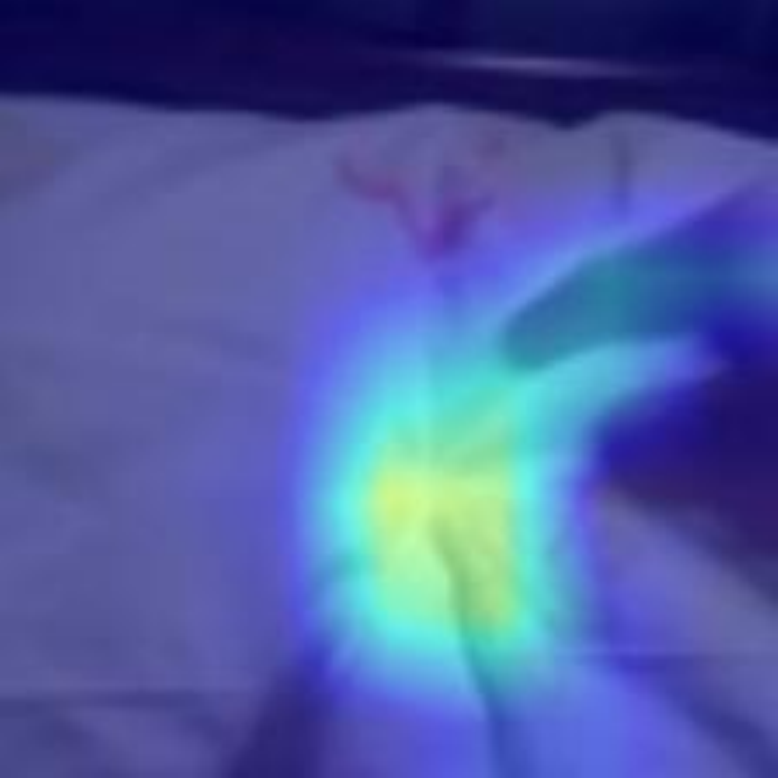}
    \end{subfigure}%
    \begin{subfigure}{.08\textwidth}
    \centering
    \includegraphics[width=\linewidth]{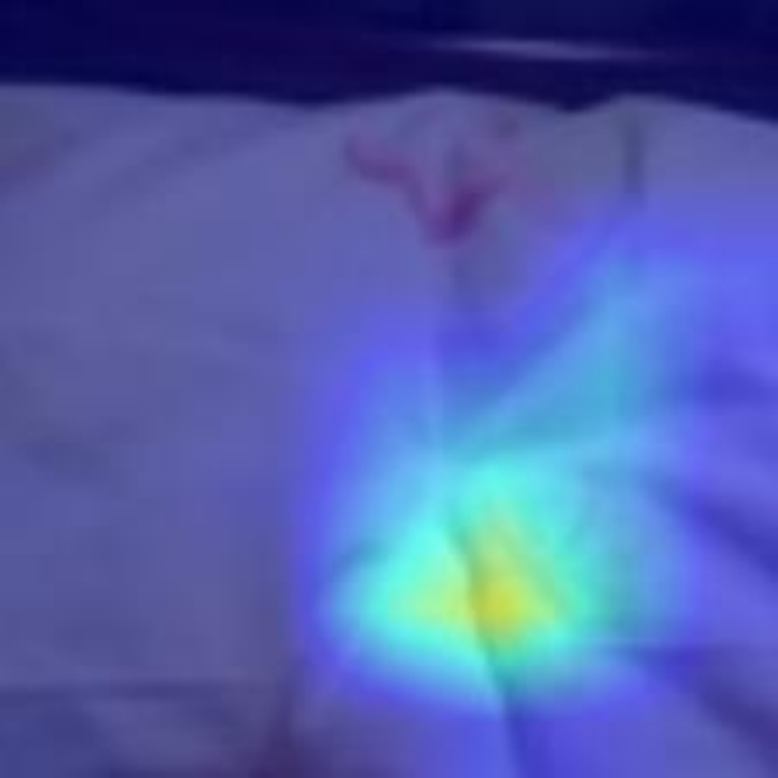}
    \end{subfigure}%

    \begin{subfigure}{.08\textwidth}
    \centering
    \includegraphics[width=\linewidth]{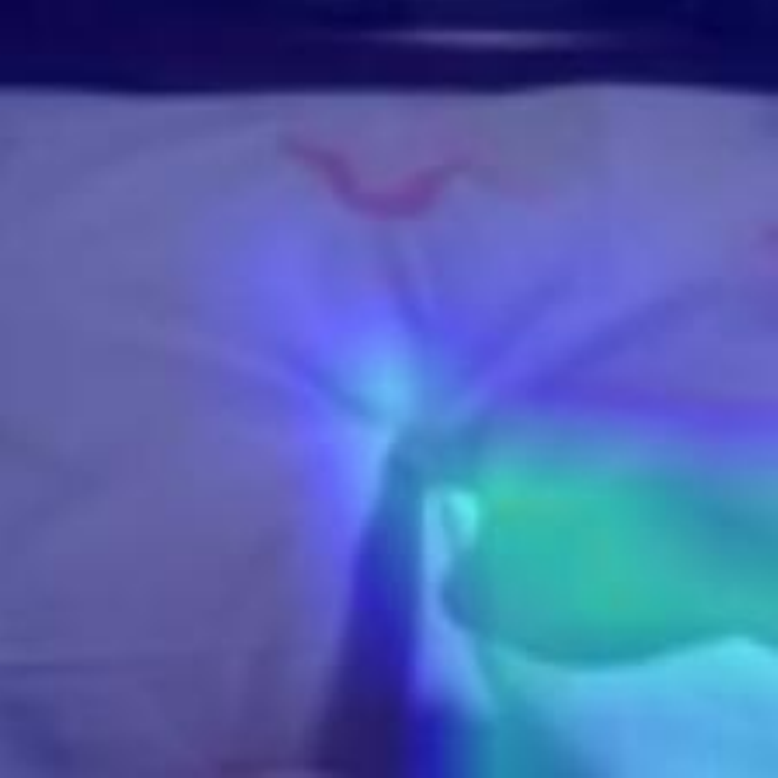}
    \end{subfigure}%
    \begin{subfigure}{.08\textwidth}
    \centering
    \includegraphics[width=\linewidth]{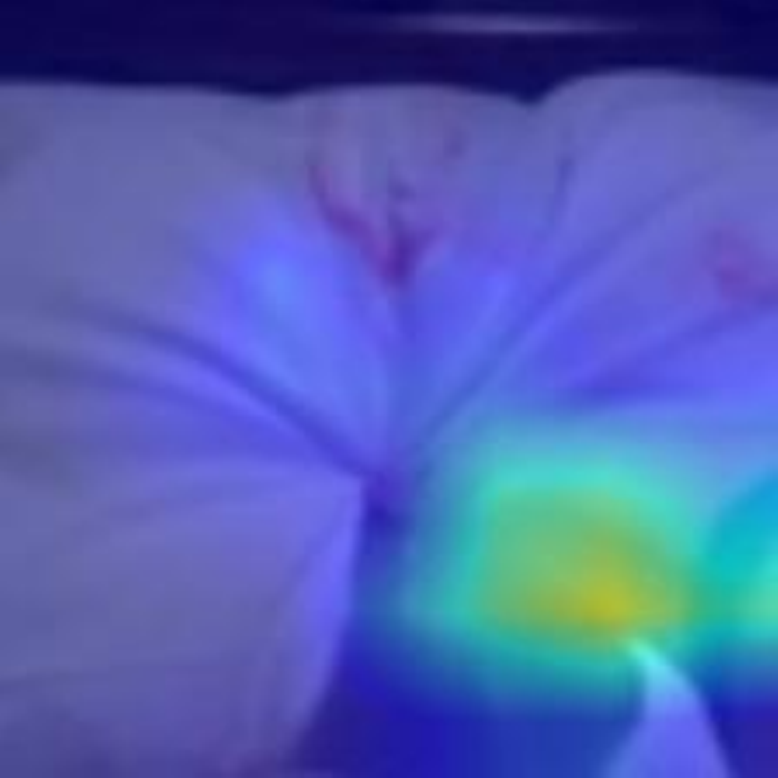}
    \end{subfigure}%
    \begin{subfigure}{.08\textwidth}
    \centering
    \includegraphics[width=\linewidth]{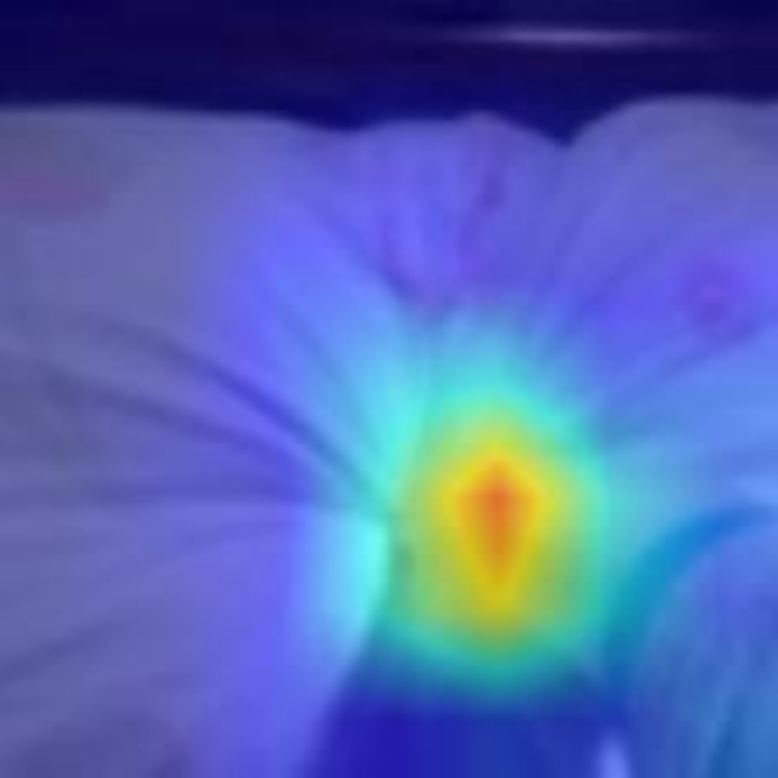}
    \end{subfigure}%
    \begin{subfigure}{.08\textwidth}
    \centering
    \includegraphics[width=\linewidth]{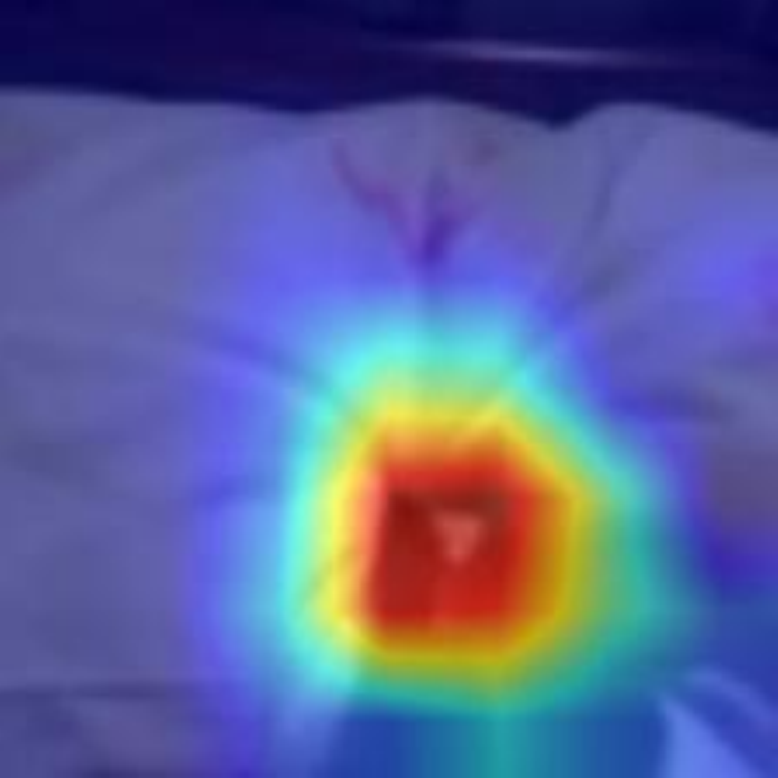}
    \end{subfigure}%
    \begin{subfigure}{.08\textwidth}
    \centering
    \includegraphics[width=\linewidth]{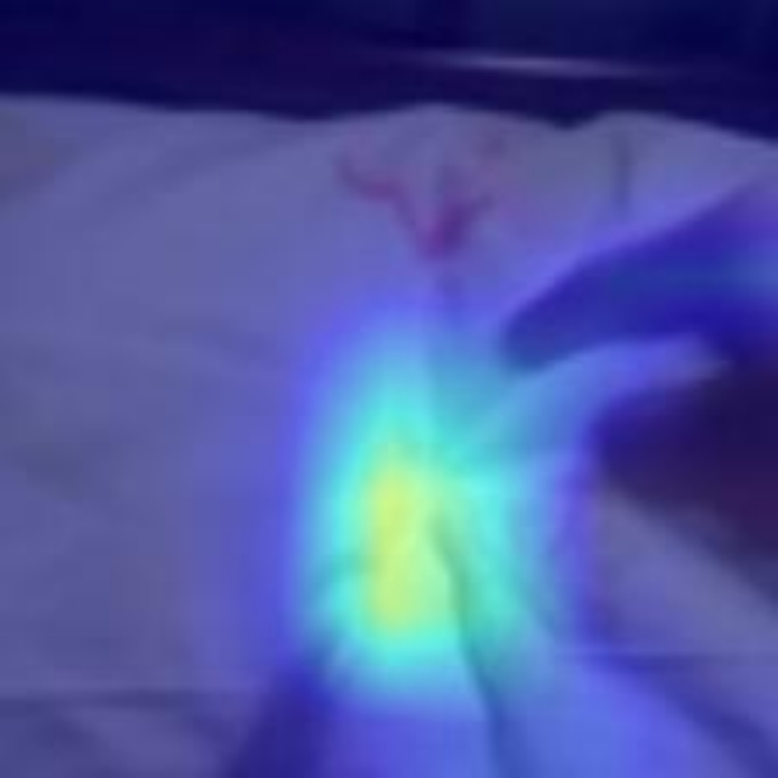}
    \end{subfigure}%
    \begin{subfigure}{.08\textwidth}
    \centering
    \includegraphics[width=\linewidth]{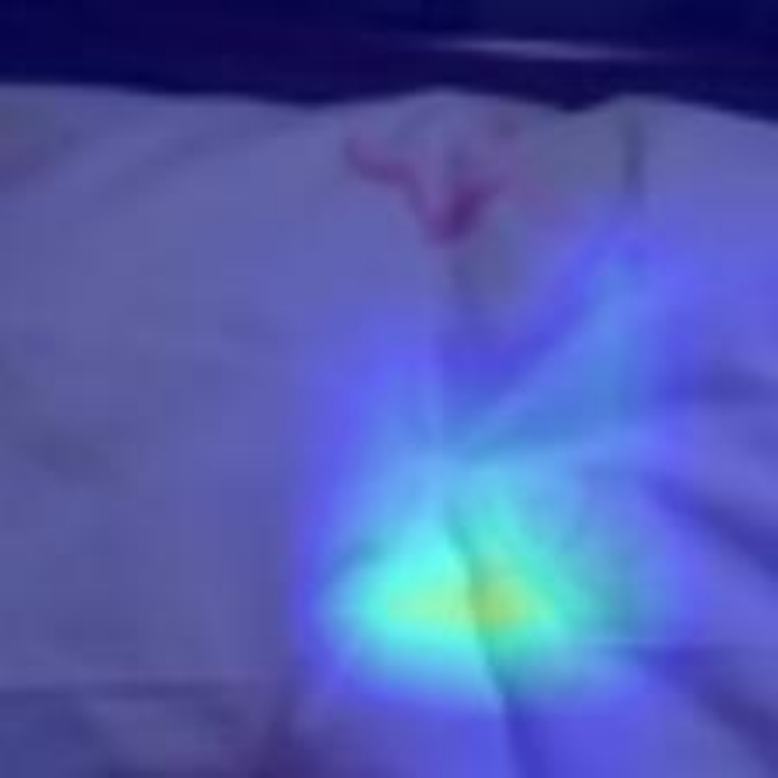}
    \end{subfigure}%

    \smallskip  
    
    \begin{subfigure}{.08\textwidth}
    \centering
    \includegraphics[width=\linewidth]{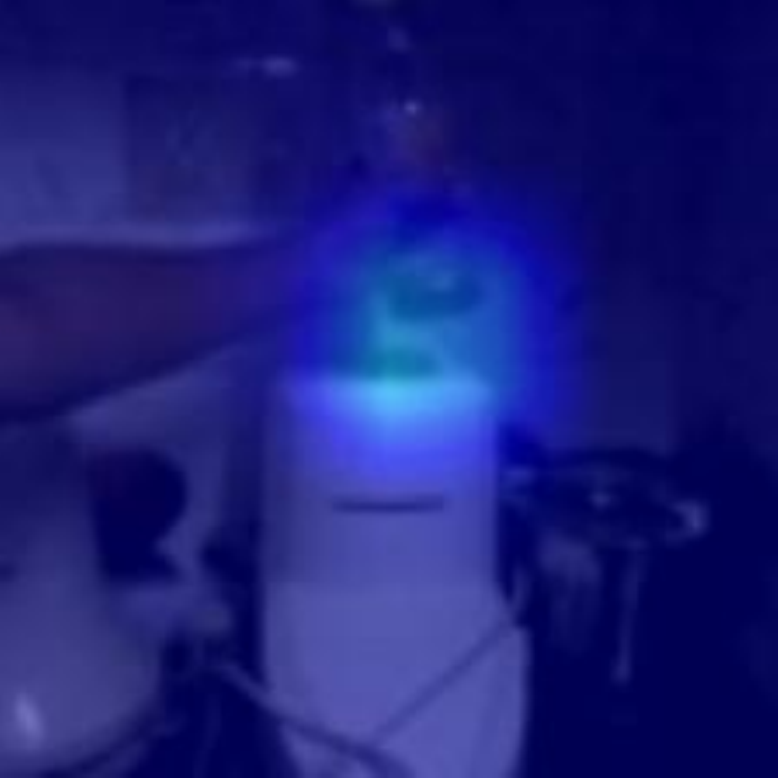}
    \end{subfigure}%
    \begin{subfigure}{.08\textwidth}
    \centering
    \includegraphics[width=\linewidth]{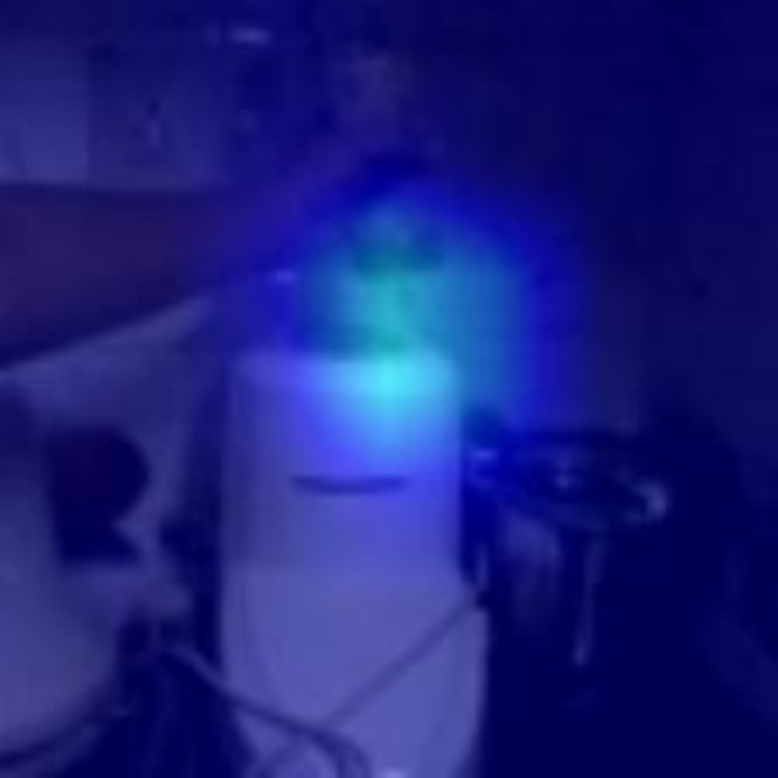}
    \end{subfigure}%
    \begin{subfigure}{.08\textwidth}
    \centering
    \includegraphics[width=\linewidth]{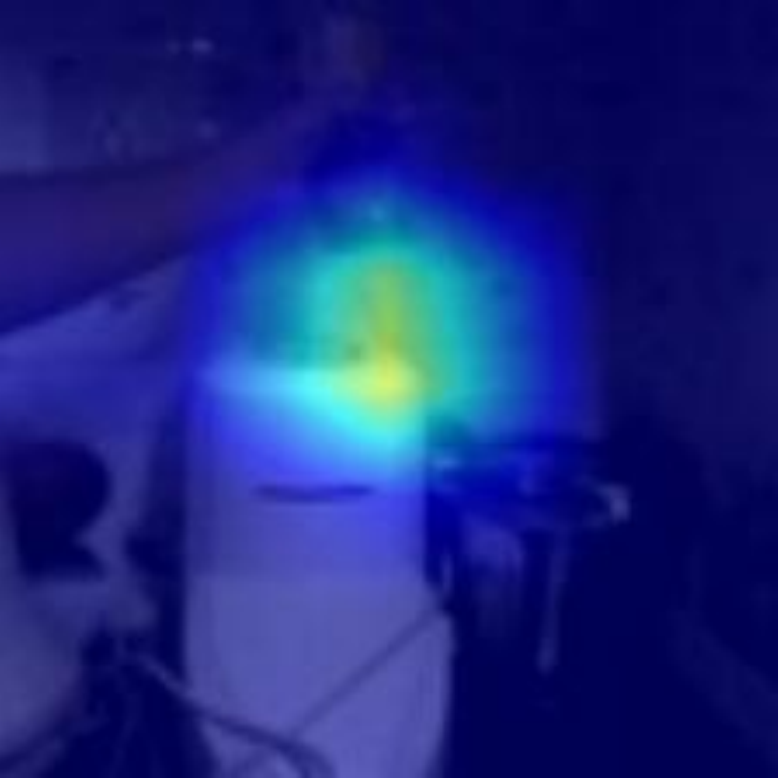}
    \end{subfigure}%
    \begin{subfigure}{.08\textwidth}
    \centering
    \includegraphics[width=\linewidth]{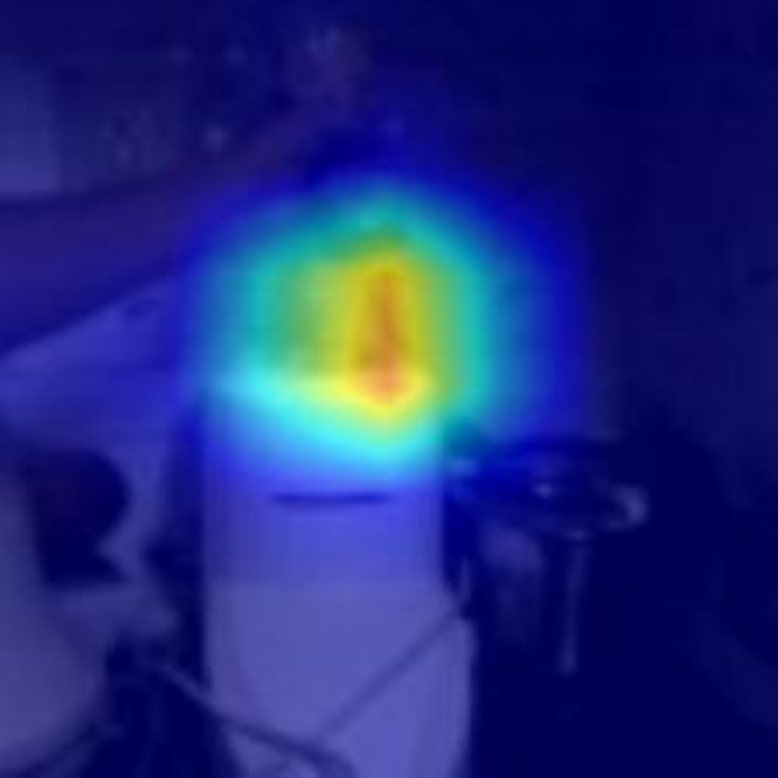}
    \end{subfigure}%
    \begin{subfigure}{.08\textwidth}
    \centering
    \includegraphics[width=\linewidth]{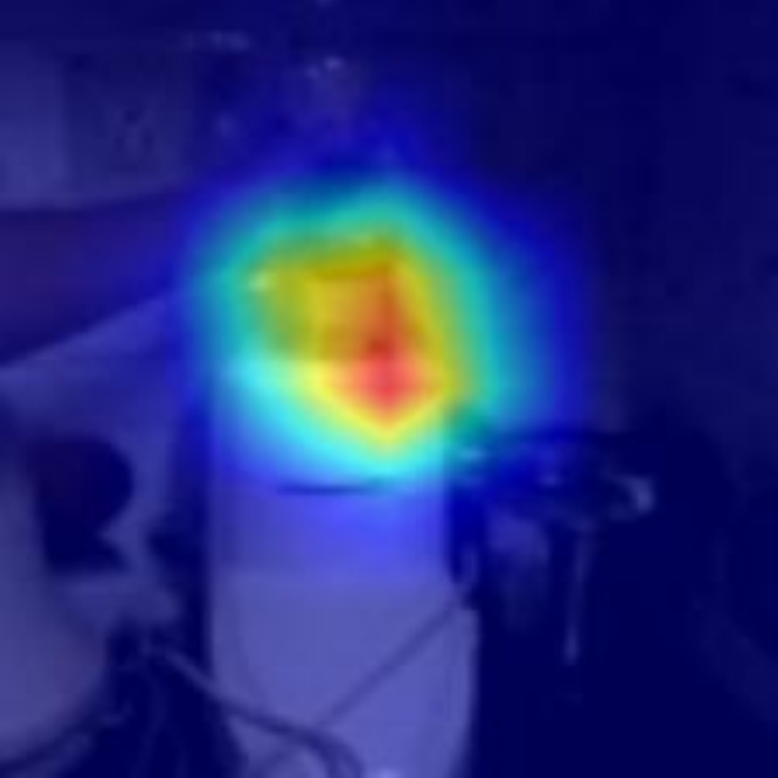}
    \end{subfigure}%
    \begin{subfigure}{.08\textwidth}
    \centering
    \includegraphics[width=\linewidth]{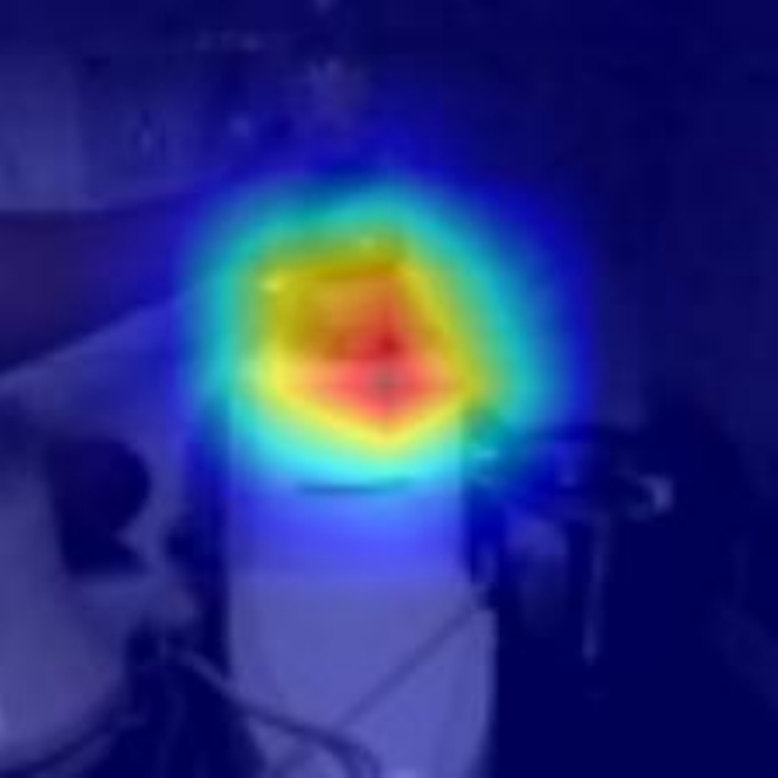}
    \end{subfigure}%

    \begin{subfigure}{.08\textwidth}
    \centering
    \includegraphics[width=\linewidth]{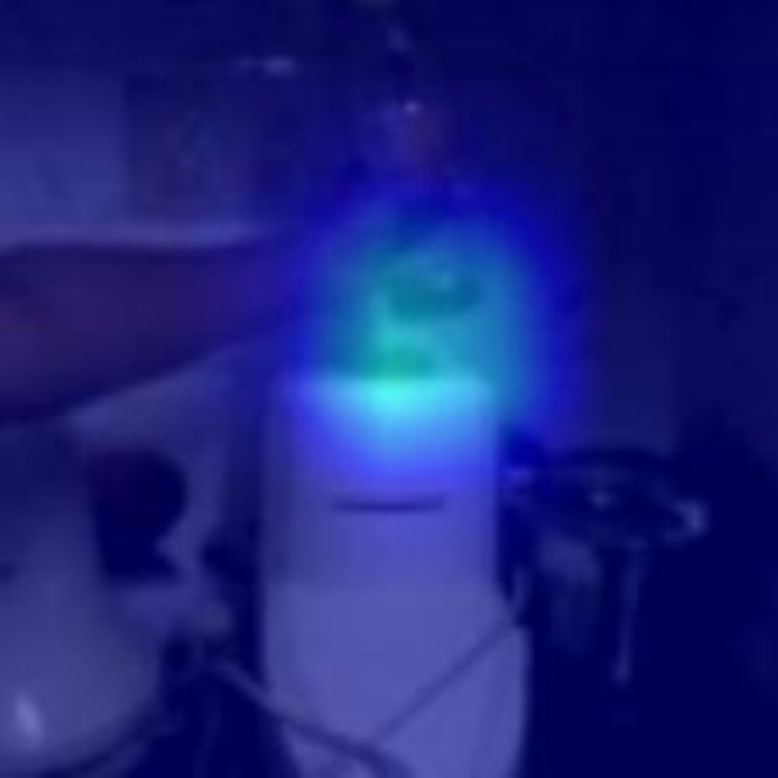}
    \end{subfigure}%
    \begin{subfigure}{.08\textwidth}
    \centering
    \includegraphics[width=\linewidth]{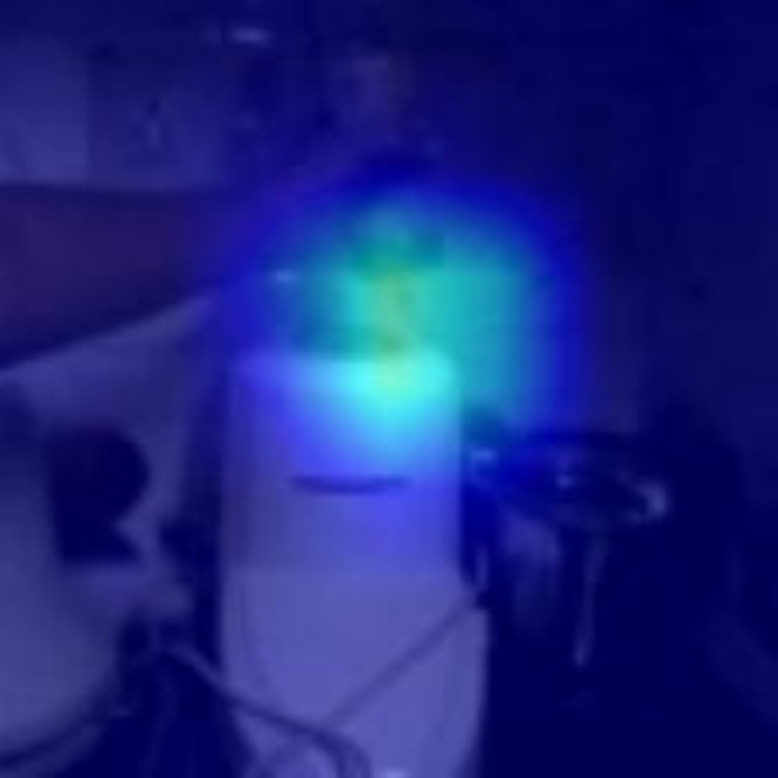}
    \end{subfigure}%
    \begin{subfigure}{.08\textwidth}
    \centering
    \includegraphics[width=\linewidth]{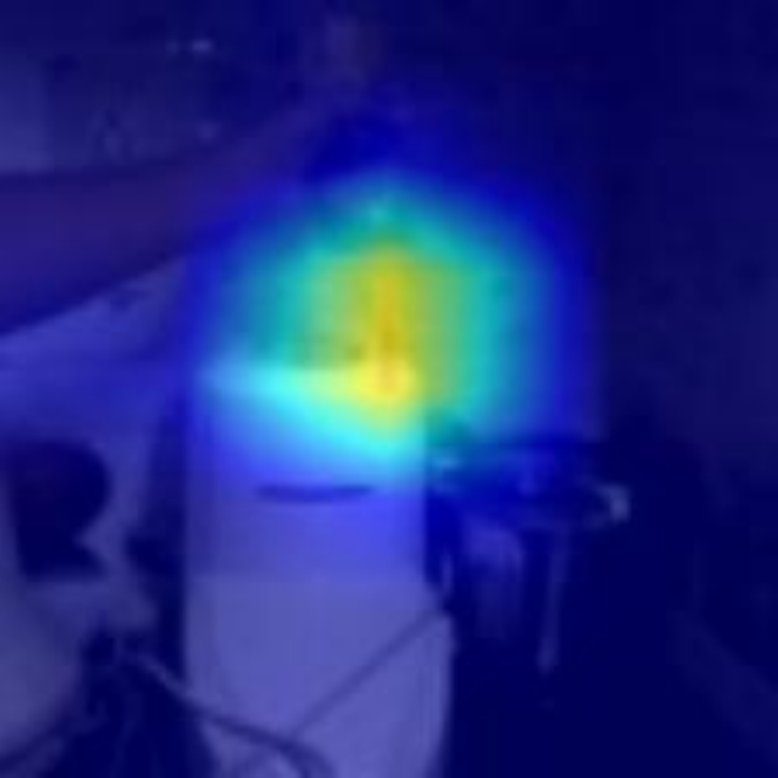}
    \end{subfigure}%
    \begin{subfigure}{.08\textwidth}
    \centering
    \includegraphics[width=\linewidth]{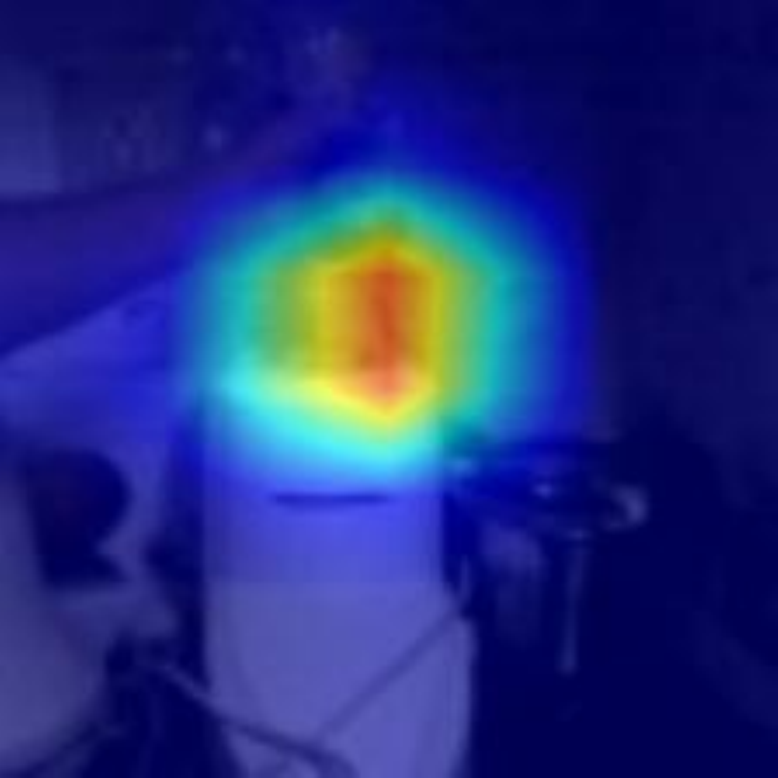}
    \end{subfigure}%
    \begin{subfigure}{.08\textwidth}
    \centering
    \includegraphics[width=\linewidth]{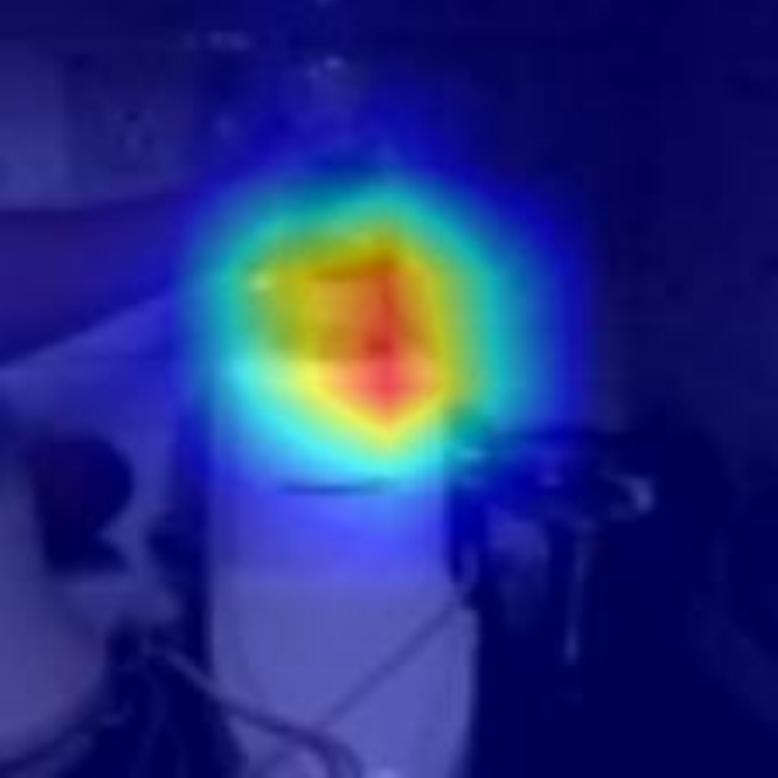}
    \end{subfigure}%
    \begin{subfigure}{0.08\textwidth}
    \centering
    \includegraphics[width=\linewidth]{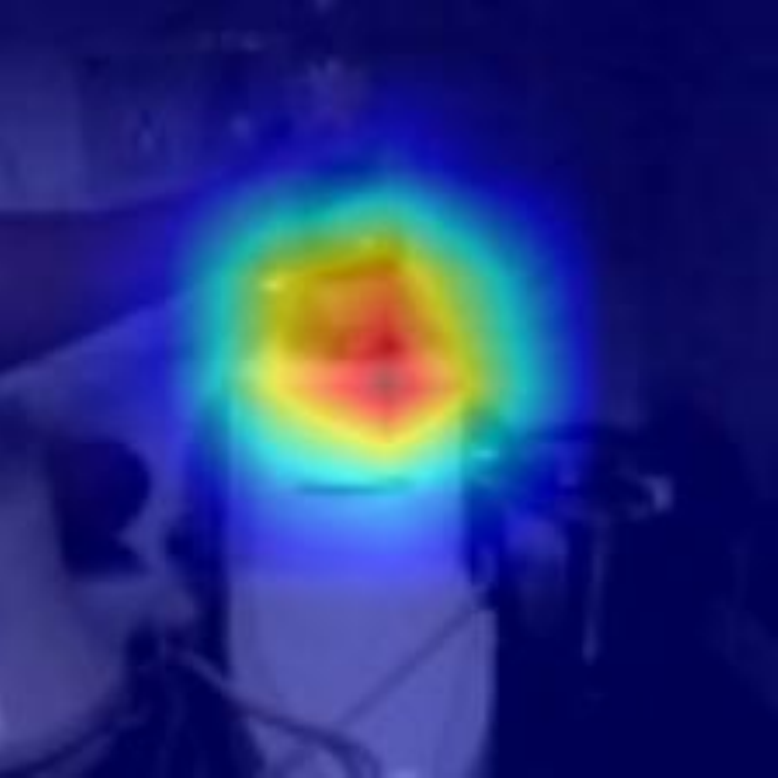}
    \end{subfigure}%

    \caption{Two representative visualizations. For each group, row 1 refers to the CAM generated by the baseline model. Row 2 represents the CAM created by model fine-tuned with TAF. TAF generates more balanced CAMs.}
    \label{fig:visualization}
\end{figure}

\subsubsection{Relieving Over-fitting}\label{sec:regularization}
We provide the training and testing curves of TSM w/ and w/o TAF in Figure~\ref{fig:curves}.
It is shown that the testing losses of TSM are significantly reduced (i.e., from approximately 2.4 to 2.2) while the training losses increases slightly during the initial fine-tuning phase, when TAF is utilized. This demonstrates that TAF effectively alleviate the overfitting issue.


\subsubsection{Computational Cost}
For training costs, since we only fine-tune models 15 epochs with very limited attack steps (i.e., $K$=1 or 3), the computational overheads are marginal. For instance, TAF will bring 25\% and 15\% additional computational costs for TSM and TPN, respectively. For inference, TAF is only applied to the training stage, the inference costs will be identical to the baseline models. 
    
\subsection{Visualization}
To qualitatively analyze the effect of TAF, a set of representative visualizations are provided in Figure~\ref{fig:visualization} and Appendix~\ref{appendix:visualization}.
Each group of visualizations includes a misclassified example and its corresponding CAM generated by the baseline model (\textbf{Row 1}), as well as the CAM generated by the model fine-tuned with TAF (\textbf{Row 2}).
Generally, CAMs from TAF and baseline model share similar trends, which is expected as TAF only fine-tunes the model for a limited number of epochs and is not expected to fundamentally change the underlying model. However, it is clear that the CAMs generated by our TAF-fine-tuned models are broader and more uniform compared to the primitive results. This observation further verifies our hypothesis that fine-tuning models with TAF can regularize temporal modeling and achieve wider attention distributions.

\subsection{Robustness Analysis}
Spurred by~\cite{hendrycks2018benchmarking}, we conduct experiments to evaluate the performance of TAF under (Out-of-distribution) OOD settings.
Three types of noise (i.e., Gaussian, Impulse, Speckle), three types of blur (i.e., Gaussian, Defocus, Zoom), and two kinds of weather corruption (i.e., Snow and Bright) are adopted.
Results are presented in Table~\ref{tab:natural corruption}.
TPN fine-tuned with TAF outperforms the vanilla TPN by 1.5\% on Weather corruptions and 1\% on other noises or blurs.
Similar results are also observed on TSM.
It reflects that the proposed TAF not only benefits the generalization of NNs but also strengthens the robustness against natural perturbations from the physical world.
The evaluations of other models in resisting natural corruption are provided in Appendix~\ref{appendix:robustness analysis}.

\section{Conclusions}
In this work, we propose TAF, a Temporal Augmentation Framework, to regularize temporal attention distributions and improve generalization in video understanding tasks. TAF leverages specifically designed temporal adversarial augmentation during fine-tuning to enhance the performance of models. Our experiments on three challenging benchmarks using four powerful models demonstrate the improvements of TAF are multi-faceted: improving video representation, relieving over-fitting issue, strengthing OOD robustness.

To the best of our knowledge, this is the first work to enhance video understanding tasks with the help of adversarial machine learning. We believe that we have established a novel and practical connection between the field of adversarial machine learning and the video understanding community.



\bibliographystyle{named}
\bibliography{main}


\appendix
\section*{Appendix}

\section{Procedures of TAF}\label{appendix:pseudo-code}
The pseudo-code of TAF is presented in Alg.~\ref{alg:TAF}.

\vspace{-3mm}
\begin{algorithm}
\SetKwInOut{KwInput}{Input}
\SetKwInOut{KwOutput}{Output}
\SetKw{KwIn}{in}
\SetKwComment{Comment}{\#}{}
\SetKwFunction{Clip}{clip}
\SetKwFunction{Sgn}{sgn}
\caption{Pseudo-code of TAF}\label{alg:TAF}
\KwInput{Dataset $D$; baseline model $\gF$ with learnable parameter $\theta$; 
        Cross-entropy loss $\gL_{ce}$; CAM operation $\gG^{C}$; 
        CAM-based loss $\gL_{\gC}$; perturbation constraint $\epsilon$; attack iterations $T$; 
        number of selected non-key frames $N$; coefficient $\alpha$; switch functions $\gS_{clean}$ and $\gS_{adv.}$.
        }
\KwOutput{Fine-tuned model $\gF$.}
\For{ $b\gets (X,y)$ \KwIn $D$}{
\small
\Comment{\,\,\,Forward through clean path.}
$S_{clean}(\gF)$

$loss_{clean} = \gL_{ce}(\theta, X, y)$

\small
\Comment{\,\,\,Generate temporally augmented examples.}
    \For{$k\gets$ 1 to $K$}{
        
        \small
        $X^{k+1} = X^{k} + \Sgn(\nabla_{X^{k}} \frac{1}{N} \sum^{N} \gG^{\gC}(\theta, X^k, \hat{y}))$
        
        $X^{k+1} = \Clip(X^{k+1}, \epsilon)$
    }
    \small
    \Comment{\,\,\,Forward through adversarial path.}
    $S_{adv.}(\gF)$
    
    $loss_{adv.} = \gL_{ce}(\theta, X, y)$
    
    \small
    \Comment{\,\,\,Update weights by jointly optimizing $\gL_{ce}$ and $\gL_{\gC}$.}
    $\gL = \alpha * \gL_{ce}(\theta, X, y) + (1-\alpha * \gL_{\gC}(\theta, X, y))$
    
    $\theta^{'} = \theta - \eta \nabla_{\theta} \gL $
}
\end{algorithm}

\section{Dataset Introduction}\label{appendix:dataset_intro}
{\bf Something-something V1\&V2} are challenging action recognition benchmarks.
There are 108,499 video clips (86,017 videos for training, 11,522 videos for validation) on V1 and 220,847 videos (168,913 for training, 24,777 for validation) on V2, with 174 categories.
Each sample in this benchmark is assigned with a specific motion label containing a placeholder, ``something'', referring to the moving object, such as ``moving something from left to right''.


{\bf Diving48} is a fine-grained temporal action recognition dataset, focusing on short-term dive actions.
It consists of 48 dive classes, with 15,027 videos in the training set and 1,970 videos for testing.
Most of the videos in Diving48 adopt the unified scene (i.e., diving competition scenes) as the motion backgrounds.
Therefore no representation biases will be introduced in this dataset.

\begin{table}[ht]
    \centering
    \begin{tabular}{lc}
        \toprule
        Method & Top-1 (\%)  \\
        \midrule
        baseline & 45.6 \\
        \midrule
        Random Noise & 45.9\\
        MixUp & 46.3 \\
        CutMix & 45.9 \\
        FrameMixUp~\cite{kim2020learning} & 43.4 \\
        CubeMixUp~\cite{kim2020learning} & 43.2 \\
        \midrule
        TAF & \textbf{46.9} \\
         \bottomrule
    \end{tabular}
    \caption{Comparison with other data augmentation methods.}
    \label{tab:comparison_aug}
\end{table}

\section{Long-term motions}\label{appendix:long term motions}

The complete list of long-term motions are summarized in Table~\ref{tab:long-term motions list}.
It is shown that 26 out of 35 long-term motions are augmented by TAF.

\section{Additional Ablation Study}\label{appendix:ablation study}

\subsubsection{Number of Attacked Frames.}
As we mentioned in Sec.~\ref{sec:temporal adversarial examples}, we choose to start attacking frames with lower CAM values.
In this part, we conduct experiments examining the importance of each frame.
Table~\ref{tab:N} shows that perturbing frames with fewer CAM values (i.e., $N$ = 2) is more efficient than attacking other frames (i.e., $N$ = 4 or 8).
Considering baseline models intend to shrink the attention scope and ignore the temporal cues from surrounding regions, shifting non-key frames causes substantial influences on the temporal receptive fields, which is consistent with our assumption.

\subsubsection{Compositions of Training Batches.}
We assign different kinds of labels $\hat{y}$ to samples based on their predictions.
As shown in Table~\ref{tab:CandW}, for TSM, training only with misclassified samples is slightly better than with correctly classified samples.
This means that reducing the difficulties of hard examples by temporal adversarial examples is more effective.
However, it is worth noting that this phenomenon is not always applicable to other situations.
For instance, models enhanced by TAF on Diving48 are trained with all the examples (i.e., $+$C, $+$W).

\subsubsection{Comparisons with other augmentations}
We provide the comparison with other popular augmentation methods in Table~\ref{tab:comparison_aug}. For random noise, as suggested, we only replace $\gamma$ with Gaussian Noise and keep other settings the same as TAF. For methods from [2], we follow their optimal settings. The results show that other augmentations cannot improve video representations and recognition accuracy as much as TAF.

\section{Visualization}\label{appendix:visualization}
Additional visualizations on rectified examples are presented in Fig.~\ref{fig:visualization in appendix}.
It is shown that TAF effectively expands the temporal attention distributions.

\section{Robustness Analysis}\label{appendix:robustness analysis}
Results of robustness analysis on Diving48 are shown in Table~\ref{tab:natural corruption appendix}.
TAF significantly boost baseline model's ability against natural corruption.
Specifically, TAF improves TSM over 3\% top-1 accuracy on resisting speckle noises and bright corruption.
In terms of other corruption, 1.5\% $\sim$ 2\% improvements are observed.
Considering the improvements on naturally corrupted dataset are remarkably larger than improvement on clean version (0.6\% in Table~\ref{tab:main_result}), it proves that TAF can effectively defense various of natural perturbations.

\section{Dataset Bias}
TAF augments action recognition models by expanding the temporal attention distributions.
It utilizes temporal adversarial examples to regularize the time-series modeling process and relieve the overfitting issues.
Therefore, TAF is greatly competitive on temporal related benchmarks.
The reason why we do not conduct experiments on scene related datasets, such as Kinetics and UCF-101, is that this kind of datasets rely less on temporal information.
Models are capable of generating precise predictions according to the texture of scenes and objects, instead of temporal cues, on these datasets.
For instance, Table 1 in~\cite{xie2018rethinking} demonstrates that \textbf{randomly} shifting the sequence of input frames does not hurt performances of models on Kinetics but will largely affect the accuracy of models trained on Something-something datasets.
This insensitivity of temporal information reduces the applicability of TAF.

\begin{table*}
    
    \centering
    \resizebox{\linewidth}{!}{%
    \begin{tabular}{l|l|c}
    \hline
index & long-term motions & $\Delta$ (\%)  \\ \hline
1 & Pretending to pour something out of something, but something is empty & 25.8 \\
2 & Poking something so that it spins around & 25.0 \\
3 & Lifting up one end of something without letting it drop down & 22.7 \\
5 & Lifting a surface with something on it but not enough for it to slide down & 16.7 \\
7 & Lifting something up completely without letting it drop down & 16.0 \\
10 & Poking a stack of something without the stack collapsing & 14.3 \\
11 & Trying to bend something unbendable so nothing happens & 14.0 \\
12 & Moving something across a surface without it falling down & 13.9 \\
14 & Pretending to close something without actually closing it & 13.0 \\
18 & Putting something onto a slanted surface but it doesn't glide down & 11.8 \\
26 & Poking a stack of something so the stack collapses & 10.0 \\
28 & Putting something onto something else that cannot support it so it falls down & 9.5 \\
31 & Lifting a surface with something on it until it starts sliding down & 8.7 \\
33 & Poking something so it slightly moves & 8.3 \\
36 & Putting something on the edge of something so it is not supported and falls down & 7.8 \\
37 & Removing something, revealing something behind & 7.8 \\
42 & Pretending or trying and failing to twist something & 6.7 \\
44 & Something colliding with something and both come to a halt & 6.7 \\
48 & Poking something so that it falls over & 6.2 \\
57 & Throwing something in the air and catching it & 5.6 \\
63 & Tilting something with something on it slightly so it doesn't fall down & 4.9 \\
78 & Trying to pour something into something, but missing so it spills next to it & 3.4 \\
80 & Putting something that can't roll onto a slanted surface, so it slides down & 3.3 \\
84 & Letting something roll up a slanted surface, so it rolls back down & 3.2 \\
91 & Pouring something into something until it overflows & 3.0 \\
107 & Throwing something in the air and letting it fall & 2.4 \\
142 & Lifting something up completely, then letting it drop down & 0.0 \\
143 & Lifting up one end of something, then letting it drop down & 0.0 \\
161 & Pushing something so that it slightly moves & 0.0 \\
163 & Putting something on a flat surface without letting it roll & 0.0 \\
165 & Putting something that can't roll onto a slanted surface, so it stays where it is & 0.0 \\
166 & Putting something that cannot actually stand upright upright on the table, 
so it falls on its side & 0.0 \\
169 & Tilting something with something on it until it falls off & 0.0 \\
170 & Tipping something with something in it over, so something in it falls out & 0.0 \\
172 & Turning the camera upwards while filming something & 0.0 \\ \hline
    \end{tabular}
    }
    \caption{Long-term motions and improvements. Motions in Fig.~\ref{fig:long-term motions} and this table share the same indexes.}
    \label{tab:long-term motions list}
\end{table*}

\begin{figure*}
    \centering
    \includegraphics[width=0.65\linewidth]{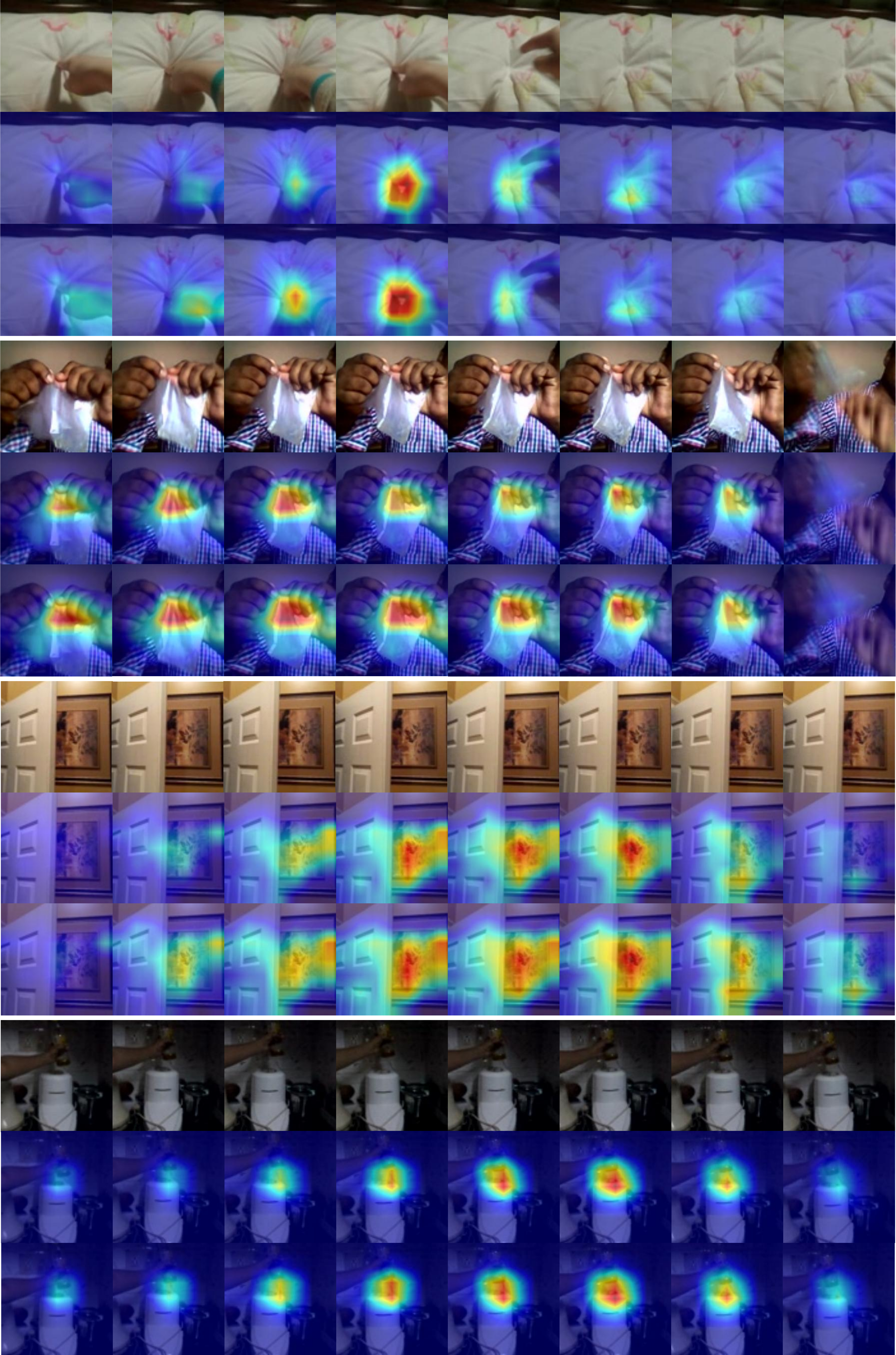}
    \caption{Visualizations on rectified samples. For each group, Row 1: natural examples; Row 2: CAMs generated by baseline models; Row 3: CAMs generated by TAF fine-tuned models.}
    \label{fig:visualization in appendix}
\end{figure*}

\begin{table*}[ht]
        
        \begin{subtable}[t]{0.4\linewidth}
        
        \centering
        \begin{tabular}{ccc}\hline
		                &   Top-1 (\%)          & Top-5 (\%) \\ \hline
		baseline        & 45.6                  & 74.6      \\ \hline
		$N$ = 2         &     46.5              &     74.8      \\
		$N$ = 4         &     46.6              &     75.0      \\
		$N$ = 8         &     \textbf{46.9}     &     75.0      \\ \hline
		\end{tabular}
		\caption{Impacts of the number of attacked non-key frames. Larger $N$ means attacking more.}
        \label{tab:N}
        \end{subtable}%
        \hfil
        \begin{subtable}[t]{0.57\linewidth}
        
        \centering
            \begin{tabular}{ccc}\hline
		                                        &   Top-1 (\%)          & Top-5 (\%) \\ \hline
		baseline                                & 45.6                  & 74.6      \\ \hline
		$+$C, $+$W          &     46.6              &     74.8      \\
		$+$C, $-$W       &     46.6              &     74.9      \\
		$-$C, $+$W       &     \textbf{46.9}     &     75.0      \\ \hline
		\end{tabular}
		\caption{Adopted samples when adversarial training. $+$/$-$ means with/without. C/W represents correctly/incorrectly classified samples.}
        \label{tab:CandW}
        \end{subtable}%
        
        \caption{Ablation study on the number of attacked frames $N$ and the compositions of adversarial training batches.}
\end{table*}

\begin{table*}
\setlength{\tabcolsep}{4pt}
    
    \centering
    \begin{tabular}{l|c|c|c|c|c|c|c|c}\toprule
              &  \multicolumn{3}{|c|}{Noise} & \multicolumn{3}{|c|}{Blur} & \multicolumn{2}{|c}{Weather}   \\ \hline
		Model &  Gauss. & Impulse & Speckle & Gauss. & Defocus & Zoom & Snow & Bright \\ \hline
		TSM   & 24.1/51.6 & 23.8/57.4 & 23.5/51.9 & 22.8/50.6 & 18.4/44.8 & 18.6/45.5 & 23.4/52.6 & 21.4/51.5 \\ \
		TSM+TAF  & \textbf{26.5}/56.9 & \textbf{25.2}/61.3 & \textbf{26.8}/56.5 & \textbf{24.3}/54.6 & \textbf{20.0}/48.3 & \textbf{20.0}/49.8 & \textbf{27.1}/56.5 & \textbf{24.9}/56.0  \\ \bottomrule
		\end{tabular}
\caption{Evaluations of defending natural corruption on Diving48 dataset. Performances are reported as Top-1(\%)/Top-5(\%).}
    \label{tab:natural corruption appendix}
\end{table*}




\end{document}